%% file: main.tex
\documentclass{article}
\usepackage[utf8]{inputenc}
\usepackage{amsmath}
\usepackage{graphicx}
\usepackage{amssymb}
\usepackage{colortbl}
\usepackage{longtable}
\usepackage{arydshln}
\usepackage{stfloats}
\usepackage{algorithm}
\usepackage{algorithmicx}
\usepackage{algpseudocode}
\usepackage{multirow}
\usepackage{extarrows}
\usepackage{mathrsfs}
\usepackage[sort]{cite}
\usepackage{hyperref}
\usepackage{array}
\usepackage{endnotes}
\usepackage{siunitx}
\usepackage{caption}
\usepackage [autostyle, english = american]{csquotes}
\MakeOuterQuote{"}

\newtheorem{definition}{Definition}

\begin{document}

\title{Verification for Machine Learning, Autonomy, and Neural Networks Survey}
\author{Weiming Xiang \and Patrick Musau \and  Ayana A. Wild \and Diego Manzanas Lopez \and Nathaniel Hamilton \and Xiaodong Yang \and Joel Rosenfeld \and Taylor T. Johnson}
\maketitle

\begin{abstract}
This survey presents an overview of verification techniques for autonomous systems, with a focus on safety-critical autonomous cyber-physical systems (CPS) and subcomponents thereof. Autonomy in CPS is enabling by recent advances in artificial intelligence (AI) and machine learning (ML) through approaches such as deep neural networks (DNNs), embedded in so-called learning enabled components (LECs) that accomplish tasks from classification to control. Recently, the formal methods and formal verification community has developed methods to characterize behaviors in these LECs with eventual goals of formally verifying specifications for LECs, and this article presents a survey of many of these recent approaches.
\end{abstract}

\input{motivation}

\input{preliminaries}

\input{application}

\input{architecture}

\input{control}

\input{learning}

\input{verification}

\input{conclusion}

\bibliographystyle{plain} 
\bibliography{ref,joelrefs}
\theendnotes
\end{document}

%% file: motivation.tex
\section{Motivation and Prevalence}
Artificial intelligence (AI) is in a renaissance, and AI methods, such as machine learning (ML), are now at a level of accuracy and performance to be competitive or better than humans for many tasks.
Deep neural networks (DNNs) in particular are increasingly effective at recognition and classification tasks.
For instance, much of the sensing, estimation, and fusion of such data that enables applications such as autonomous driving and other autonomous cyber-physical systems (CPS) increasingly relies on DNNs and similar ML techniques.
However, this progress comes at significant risk when these methods are deployed in operational safety-critical systems, especially those without direct human supervision.

Building on applications of AI/ML in autonomous CPS, this paper surveys the current state-of-the-art for safely integrating AI/ML components---which we term \emph{learning enabled components (LECs)} into safety-critical CPS.
This survey consists of several major parts, focused around how machine learning is enabling autonomy in CPS, including: illustrative applications, intelligent control, safety architectures with AI/ML components, understanding AI/ML components and systems through statistical and symbolic methods---such as specification inference and automata learning---safe planning, reachability analysis of AI/ML components, such as neural networks, etc.


%
Much of the recent progress in AI/ML relies on advances in statistical methods, such as DNNs, where it is difficult to understand how components operate and how decisions are made.
In contrast, much recent progress in formal methods relies around symbolic methods, where semantics and operational behaviors are precisely specified.


%% file: preliminaries.tex



%% file: application.tex
\section{Applications: Autonomous Driving}
Recent advances in machine learning paradigms have stimulated significant progress in the development of autonomous driving applications. Autonomous driving can be split into two major concepts: mediated perception approaches and behavior reflex approaches \cite{ChenDeepDriving}. Mediated perception regimes adopt numerous sub-components for identifying driving related objects, such as traffic lights, lanes, signs, and pedestrians, in order to create a representation of the car's surroundings. Using this information an AI based engine is utilized in order to identify the relevant objects in order to correctly control the car's velocity and direction\cite{ChenDeepDriving}. Behavior reflex approaches create mappings from sensory input images to driving actions. In these schemes, a neural network is trained to mimic a human driver by recording steering angles and sensory images recorded on a series of driving runs. While these schemes have proved to be largely successful, there is an urgent need to formally reason about the behavior of autonomous driving systems due to their safety critical nature. As a result, there has been great research impetus towards obtaining robust and scalable verification schemes. Thus, in this section, we present a brief survey of the existing verification approaches for autonomous driving applications.

The paper \cite{wrro127573} describes the purpose and  progress of the Safety of Autonomous Systems Working Group (SASWG), which was established by the Safety Critical Systems Club (SCSC). It emphasizes the importance of careful and tactful implementation of autonomous systems. They discuss the problems that are close to being solved, the problems with unclear solutions, and the problems with possibly no solution. The approach described in the paper is for the problem of autonomous systems safety in general.

In \cite{2018arXiv180406760E}, the authors present a testing framework for test case generation and automatic falsification methods for cyber-physical systems called Sim-ATAV (Simulation-based Adversarial Testing of Autonomous Vehicles). This framework can be used to evaluate closed-loop properties of autonomous systems that include machine learning components using a test generation method called covering arrays. This is all performed in a virtual environment where the focus is on identifying perturbations that lead to unexpected behaviors. By utilizing advanced 3D models and image environments, the authors seek to bridge the gap between the virtual world and the real world. Their work demonstrates how to use test cases in order to identify problematic test scenarios and increase the reliability of autonomous vehicles. While most current approaches focus only on the machine learning components, the authors of this paper focus on the closed loop operation of the vehicles. It deals with verification and testing at the system level. Their paper demonstrates effective ways of testing discrete combinations of parameters and corner cases.


In a recent paper~\cite{LangleyML2011}, Pat Langley details the history of the field of Machine Learning and the reasons for launching a new journal in the 1980's. Langley argues that Machine Learning was originally concerned with the development of intelligent systems that would display rich behavior on a set of complex tasks. However, in recent years, many researchers have focused on tasks that do not tackle intelligence or systems and have mainly been preoccupied by statistics and classification. This paper focuses on the need to "recover the discipline's original breadth of vision and its audacity to develop learning mechanisms that cover the full range of abilities in humans."
%
Safety for cyber-physical systems that utilize controllers of neural networks needs to be verified using approaches such as those based in simulations~\cite{DBLP:journals/corr/abs-1804-03973}.

%% file: architecture.tex
\section{Architecture: Safe Monitoring and Control}
It is unrealistic to believe that one will ever verify all parts of an autonomous CPS, as such a problem would be akin---and in fact far harder---than formally verifying that all components in a modern microprocessor consisting of billions of transistors meets a formal specification of every desirable behavior.
Aside from it being unlikely that every specification would ever be formalized for such complex systems, the computational complexity of subproblems such as model checking, equivalence checking, etc. will never be fully realized, as the complexity is typically exponential in the size of the system (under some definition of size, such as lines of source code) and additionally the size of such systems is growing exponentially, especially as LECs are incorporated.
As such, some of the most plausibly impactful methods to ensure autonomous CPS meet their specifications is through safety architectures, runtimne monitoring, runtime verification (RV), and runtime assurance (RTA).

The paper \cite{johnson2016tecs} analyzes periodically-scheduled controller subsystems of cyber-physical systems using hybrid automata and associated analysis tools. However, reachability analysis tools do not perform well on periodically scheduled models due to a combination of large discrete jumps and nondeterminism associated with the start time of a controller. Thus, in this paper, the authors propose and demonstrate a solution that is validated through examining an abstraction mechanism in which every behavior of the original sampled system is contained in the continuous system. The authors also add a nondeterministic input. Using this framework, the authors demonstrate that reachability tools can better handle systems of this nature. Their approach is automated using the Hyst model transformation tool.

In \cite{bak2015rtss}, Stanley Bak et al. present an alternate design for the Simplex Architecture that leverages linear matrix inequality optimization and hybrid systems reachability. In this framework, the region in which a complex controller can be use is extended by using real-time reachability computations. Additionally this approach decreases the conservatism in the switching logic. This work is the first to present a viable reachability algorithm based on a system's real-time conception of imprecise computation. Moreover, their algorithm returns an over-approximation of the reachable set that is improved over time. In the experimental evaluation of their methods, the authors demonstrate that a complex controller improves significantly using quick reachability calculations lasting tens of milliseconds that bound the future behavior of the system. Finally, they demonstrate that real-time reachability has applications beyond the Simplex architecture.

The paper \cite{vijayakumar2018neural} presents a hybrid synthesis technique called Neural Guided Deductive Search (NGDS) that combines symbolic logic techniques and statistical models in order to produces programs that generalize well on previously unseen examples and satisfy a set of specifications. The programs are similar to data-driven systems. In their approach, the authors utilize deductive search in order to reduce the learning problem of a neural component into a simple supervised learning set up. Thus, as a result of this reduction, the authors are able to train their model on a limited amount of real world data and leverage the power of recurrent neural network encoders. The experimental evaluation of their methods demonstrates that their approach is superior to other approaches since it guarantees correctness and adequate generalization. Their approach is also faster than other synthesis regimes.

The paper by Saunders et al~\cite{SaundersTrial2017} presents a safe reinforcement learning regime, called HIRL (human intervention reinforcement learning), that applies human oversight to the learning process in order to prevent catastrophic executions during the exploration process. In this framework, a human supervisor observes all of the agents actions and either allows an action to take place or blocks an action in the interest of safety. Once the human supervisor classifies a sufficient number of  actions, the authors train a supervised learner to carry out this supervisory role and eliminate human intervention. In the experimental evaluation of their methods, the authors demonstrate that HIRL is successful in preventing harmful actions and that the learning process executes within an acceptable time period.

%% file: control.tex
\section{Intelligent Control}

\par Intelligent control can be spilt into many sub-domains utilizing techniques like, Neural Networks (NNs), Machine Learning (ML), Bayesian probability, fuzzy logic, and neuro-fuzzy hybridization. However, this section focuses primarily on intelligent control using Neural Networks but future revisions will include the other sub-domains. Neural Networks are systems designed to mimic the human brain. Each node in a NN is capable of making simple calculations. When nodes are connected in a network, the system is capable of calculating linear and nonlinear models in a timely and distributed manner. For more in-depth information about NNs, the introduction section of \cite{170966} contains a brief history of NNs. Additionally, \cite{80202} provides a foundational knowledge of how NNs and Recurrent Neural Networks (RNNs) have been combined and are used today in the field. 

The remainder of this section consists of summaries of influential and novel papers in the field of intelligent neural network control. Each paragraph summarizes a different paper.


In \cite{DBLP:journals/corr/ZribiCD15}, an improved gradient descent method to adjust the PID Neural Network (PIDNN) parameters is used. A margin stability term is employed in the momentum function to modify the training speed depending of the robustness of the system. As a result, the learning algorithm system converges faster. The system is structured as a feedforward NN, where the PIDNN is placed in cascade with the plant NN model. The PIDNN always has one hidden layer with 3 neurons, representing the Kp, Ki, and Kd used in traditional PID controllers.

In \cite{4812095}, a mixed locally recurrent neural network is used to create a PID Neural Network (PIDNN) nonlinear adaptive controller for an uncertain multivariable  SIMO (single-input/multi-output) system. The PIDNN is placed in cascade with the plant, and the output of the plant is fed back to the PIDNN, which is subtracted from the reference input, to calculate the error that is used to update the weights of the PIDNN. The PIDNN structure studied consists of a single input and multiple outputs,  one hidden layer with 3 neurons, the integral node with an output feedback, the derivative node with an activation feedback, and the proportional node (general node). All the activation functions are linear.  The proposed controller can update the weights online using the resilient gradient descent back-propagation algorithm with sign. The only requirement for this method is that the initial weight values are required to start the system. The initial weight can run the system stably. 

In \cite{501721}, the authors focus on the stability of a control system with neural networks using a Lyapunov approach, which is derived using a parameter region (PR) method for the representation of the nonlinear parameters' location. A barrier Lyapunov function (BLF) is introduced to every step in a back-stepping procedure to overcome the effect of the full-state constraints. Fewer learning parameters are needed in the controller design. The dynamics of the neural network system are treated as linear differential inclusions (LDIs). The NN control system consists of an approximated plant and its controller, which is also represented by LDIs. Finally, they prove that all the signals of the closed-loop system are bounded and that the tracking error converges to a bounded compact set. 

In \cite{7185423}, an intelligent control methodology based on backstepping is developed for uncertain high-order systems. They show that the output-tracking performance of the system is achieved by a state feedback controller under two mild assumptions. By introducing a parameter in the derivations, the tracking error can be reduced through tuning the controller design parameters. In order to solve overparameterization, a common problem for adaptive control design, a controller with one adaptive law is proposed.

In \cite{7087381}, an adaptive NN control and Nonlinear Model Predictive Control (NMPC) approach is proposed for a nonlinear system with a double-layer architecture. At the device layer, an adaptive NN control scheme is presented to guarantee the stability of the local nonlinear plant. Combining the overall performance index and index prediction unit, formed using radial basis function NNs, an NMPC method is proposed to optimize the performance index under the effects of network-induced delays and packet dropouts.  

In \cite{1296692}, a neural network controller for nonlinear discrete-time systems is designed, based on the structure of a linear Generalized Minimum Variance (GMV) controller. The system implemented consists of a fixed linear nominal model and a recurrent neural network that updates at each sampling interval. The adaptive control algorithm is derived by combining NNs and Lyapunov synthesis techniques, guaranteeing the stability of the system under certain assumptions. This method is tested with two simulation examples to show its efficiency.

In \cite{1215406}, multilayer NNs are used to approximate the implicit desired feedback control. Under the Lipschitz condition assumption, by using converse Lyapunov theorem, they show that the system’s internal states do remain in a compact set. The main contributions are: a state feedback control and an observer-based NN output control of a SISO (Single-Input/Single-Output) nonaffine nonlinear system with zero dynamics using multilayer NNs, which can also be extended to affine systems. Finally, they proved the existence of an implicit feedback control based on the implicit function theorem.

In \cite{363441}, two Diagonal Recurrent Neural Networks (DRNNs) are used for this control system, one as the controller and the other one as an identifier of the plant, whose output is used by the controller. Both have one hidden layer with sigmoid-type recurrent neurons. This control system is compared to Feedforward Neural Network (FNN) and Feedforward Recurrent Neural Network (FRNN) control systems. The advantages of using this system are fewer weights are required than with the FRNN model and it has dynamic mapping characteristics. These features allow the system to be used for online applications. A generalized dynamic backpropagation is used for the training of the parameters of the Diagonal Recurrent NeuroIdentifier (DRNI) and Diagonal Recurrent NeuroController (DRNC). Also, to ensure stability of the plant, they take an approach to find the bounds on learning rates based on the Lyapunov equation.

In \cite{165588}, a neural network with one hidden layer of Gaussian radial basis functions is used to adaptively approximate the nonlinear dynamics of a system with a direct adaptive tracking control architecture. Considering some assumptions about the degree of smoothness of the nonlinear dynamics, the algorithm is proven to be globally stable. Using Lyapunov stability theory, a table weight adjustment mechanism is determined. Also, the tracking errors are shown to converge to a small neighborhood around zero. Another novelty of the paper is its unique feature of transitioning between adaptive and non-adaptive modes during its operation.  

In \cite{6937163}, a backstepping-based adaptive neural output-feedback control method is developed for nonstrict-feedback stochastic nonlinear systems with unmeasurable states, which most of the previous works assume are measurable. This method is based on the state observer and the combination of radial basis function and neural network's approximation capability. In this adaptive control system, only one adaptive parameter is involved, so the computational cost is reduced. This approach guarantees that the signals in the control-loop system remain semi-globally uniformly ultimately bounded and that the observer errors and output converge to a small range around the origin.

In \cite{6783745}, a composite neural Desired State Configuration (DSC) design is studied for a class of SISO strict-feedback systems. The system is composed of radial basis function NNs and their goal is to eliminate the problem of explosion complexity. The adaptive control system uses the tracking error as well as a prediction error, which is calculated based on the accuracy of the identified neural models. In the composite NNs, the updating weights laws provide the ability of fast adaptation. Utilizing the Lyapunov method, the uniformly ultimate boundedness stability is verified. 

The goal of \cite{5430947} is to design a robust control system of uncertain multiple-input-multiple-output (MIMO) nonlinear systems with input deadzone and saturation. Considering the saturation and deadzone nonlinearities, a variable structure control (VSC) is presented. The cascade property of the system is used in developing the control structure and NN weight learning laws. Looking at the case when the control coefficient matrices are unknown, a robust backstepping technique in combination with NN approximation of the nonlinearities, the VSC control technique, and a Lyapunov synthesis, a control system is proposed. This method guarantees the stability of the closed-loop system and tracking errors converge to small residual sets.

In \cite{7066958}, a decentralized control approach for MIMO large-scale systems with uncertain stochastic nonlinear dynamics and strong interconnections is proposed. The nonlinear dynamics are approximated using radial basis functions neural networks (RBFNNs), and the adaptive neural decentralized controller is constructed using backstepping technique. The other main contribution is the use of only one parameter in the control system, for each subsystem of n-order dynamics, which helps significantly improve the computational cost. It is proven that all signals are semi-globally uniformly ultimately bounded (SGUUB), and the tracking errors converge to a small range around the origin. Simulation examples are employed to show the effectiveness of this method.

In \cite{7429795}, an adaptive neural network control method is presented for a class of SISO nonlinear systems with both unknown function and the full-state constraints. The full-state constraints are overcome by using a barrier Lyapunov function in every step for the backstepping procedure and their effect is mitigated. Another improvement, with respect to previous works, is the use of less learning parameters of nonlinear systems with unknown functions and full-state constraints, which reduces the computational cost. Using a BLF, it is proven that the signals are semiglobally bounded and the errors are driven to a small neighborhood around 0.

In \cite{7087381}, a multirate networked industrial process control problem is considered using a double-layer architecture.  At the device layer, where RBFNNs are used to approximate the nonlinear functions, an adaptive NN control approach is designed to solve the tracking problem and guarantee the stability of the plant. This layer is combined with the operation layer, where an NMPC method is proposed for optimizing the performance index under the effect of a network-induced delays, which is the result of the different sampling methods used in each layer. The lifting method is used to solve the problem of different sampling sizes. Finally, a stochastic variable satisfying the Bernoulli random distribution is utilized to model the network-induced packet dropout phenomenon at the operation layer.

In \cite{6514578}, a method for learning from adaptive neural network control of a class of nonaffine nonlinear systems in uncertain dynamic environments is studied. The nonaffine nonlinear system is first converted into a semi-affine system using a filtered tracking error, the mean value theorem, and the implicit function theorem. Under input-to-state stability and the small gain theorem, the adaptive NN tracking control is designed, which is able to relax the constraint conditions of this class of systems, on top of overcoming its complex controllability problem. Under the persistent excitation (PE) condition (due to the use of RBFNNs), this proposed technique acquires knowledge from the implicit desired control input in the stable control process and stores the collected knowledge in memory. With this dynamic learning proposed, the NN learning control improves the control performance and enables closed-loop stability.

In \cite{6627983}, an adaptive tracking control algorithm for nonlinear stochastic systems with unknown functions that uses backstepping is proposed and mathematically proved to work. There are fewer adjustable parameters than other algorithms which reduces the online computation load. They use Lyaponov analysis to prove that all the signals in the system are SGUUB in probability and that the system outputs track the reference signal to a bounded compact set. Their example is given as a simulation, no real-world tests were done. Future research will extend the usage to control MIMO systems with stochastic disturbance.

This paper, \cite{ZHAO2015193}, proposes and proves an algorithm for handling switched stochastic nonlinear systems that have nonstrict-feedback with unknown nonsymmetric actuator dead-zones and arbitrary switching. They combine radial basis function neural network aproximation and adaptive backstepping with  stochastic Lyapunov methods to develop a system that is SGUUB by the 4th moment. They show their results through a siulation with a ship manuevering system, but no real-world tests were completed. Future research will try to apply the same method to more complicated systems, such as high-order switched stochastic systems.

This paper, \cite{6651788}, focuses on reducing vibrations in flexible crane systems, represented as partial-ordinary differential equations, by using an integral-barrier Lyapunov function (IBLF)-based control system. This technique is primarily used for ODE systems, but they are able to adapt it to fit their constraints. The simulated results are very promising.

In \cite{572089}, two linear control models, two different Taylor expansion approximations of the nonlinear autoregressive moving average (NARMA) model, are presented. These two models are the NARMA-L1 and NARMA-L2 models. The fact that these controllers are linear in the control input extensively simplifies both the theoretical analysis and implementation. The primary case studied is when only the input and output are known, which means that the system identification and control have to be studied using only data from the input and output. The results obtained show that these models result in better control than an exact NARMA model, since dynamic gradient methods can be computationally intensive, so an approximation of them results in better identification. 

This paper, \cite{80202}, focuses on the identification and control of nonlinear dynamic plants using neural networks. They propose that multilayer neural networks and recurrent networks should be united for future use and show simulation results proving their claims. The paper is more foundational and should be read by people just starting out in the field because their explanations are in-depth and well cited.

This paper, \cite{7086072}, presents an event-triggered control design for multi-input multi-output uncertain nonlinear continuos-time systems. By using neural network approximation properties in the context of event-based sampling, they found a way to reduce the network resource overhead using the Lyapunov technique to create an event-triggered condition. Additionally, the paper introduces a weight update law for aperiodic tuning of the neural network in order to relax the knowledge of complete system dynamics and reduce computation. Through proofs and simulations, they were able to prove closed-loop stability for the design.

This paper, \cite{6736141}, proposes an iterative two-stage dual heuristic programming (DHP) method and a nonquadratic performance functional to solve optimal control problems for a class of discrete-time switched nonlinear systems subject to actuator saturation. The nonquadratic performance functional confronts the control constraints of the saturating actuator and the DHP method solves the Hamilton-Jacobi-Bellman equation. The mathematical proofs are validated by simulation results.

This paper, \cite{7451284}, focuses on adaptive trajectory tracking control for a remotely operated vehicle (ROV) with an unknown dynamic model and an innaccurate thrust model. They propose a local recurrent neural network (local RNN) structure with a fast learning speed to hande these shortcomings. Their simulation results show that their method's tracking error converges in a finite time. However, the authors recommend future work to shorten the amount of time it takes to converge.

In \cite{1281757}, the authors exhibit an approach to analyze neural network performance in control systems that are advanced adaptive. A tool is presented for measuring neural network operational performance from the output's error bar calculation. The tool can be used prior to deploying the system or while the system is in operation. The main purpose of the tool is to further neural network reliability in systems that are adaptive. Further work could include creating tools for verification and validation of adaptive systems and apply technology for adaptive neural networks to actual missions.


In \cite{WEI2015106}, the authors present an algorithm for iterative adaptive dynamic programming (ADP) for problems of optimal tracking control on infinite horizon nonlinear system that are discrete-time. The algorithm efficiently solves the optimal tracking control problems. The performance index function's convergence analysis technique and the least upper bound of said function are developed and exhibited. They use neural networks for the iterative ADP algorithm implementation, for performing iteratively and tracking optimally.  Further work could include analyzing the stability of the algorithm, examining the properties that interact with the system, and examining the reasons for errors in the iterations.

In \cite{7067402}, the authors present a novel approach for backstepping output-feedback control of adaptive neural networks. Through the combination of radial basis functions and the aforementioned control approach, a controller for output-feedback adaptive neural networks was designed. In a closed-loop system, the controller supports boundedness, on a semiglobal level, for the signals in the system. The controller is adapted for nonstrict-feedback systems but can be applied to controlling both nonstrict- and strict-feedback systems. The control scheme can handle systems that are nonlinear and have nonstrict-feedback, making it a broad coverage scheme.

In \cite{170966}, the authors present a survey of the theory and applications for control systems of neural networks. They give an overview of neural networks and discuss the benefits of them. Neural networks, like in the brain, have parallel processing, learning, mapping that is nonlinear, and generalization capabilities. The general neural network model is discussed along with recurrent and feed-forward neural networks. Some applications of neural networks that are discussed are recognition of patterns, planning and design, control, processing of knowledge information, and systems that are hybrid. The concept and applications of control that is based on neural networks, neuromorphic control, are discussed in detail. The last topic discussed in the paper is the fusion of fuzzy technology, artificial intelligence, and neural networks to create the FAN technology. 

In \cite{1866}, Behnam Bavarian presents a neural network based approach for the design and implementation of intelligent control systems. The author explores general problems in intelligent controller design where the control problem can be formulated as a broad mapping from a series of state changes to state actions. Two examples are discussed in the paper: the first explores the use of a bidirectional associative memory neural networks for sensor and actuator fault isolation, and the second deals with a hopfield neural network used to solve the traveling salesman problem. Overall, the authors demonstrate that neural networks represent a promising avenue of achieving intelligent control.

The paper, \cite{Lewis}, investigates the promise of using neural networks for model-free learning controllers for non-linear systems. The authors primarily focus on multi-loop controllers where a neural networks is present in some of the loops and an outer unity-gain feedback loop. The authors further demonstrate that as the uncertainty about the controlled system increases, neural network controllers form an elegant control architecture that exhibit the hierarchical structure of other approaches. The topics explored in this paper include feedback linearization design of neural network controllers, neural network control for discrete time systems, feed-forward control structures of actuator compensation, reinforcement learning control using neural networks, and optimal control.

The paper, \cite{7468475}, considers the trajectory tracking of a marine surface vessel with positional constraints in the presence of uncertainties in the environment. The authors use neural networks, in an adaptive control format, to model the environmental uncertainties. Their approach makes use of output feedback control defined by the Moore-Penrose pseudoinverse and state feedback control laws using a high gain observer. The authors demonstrate their method's abilities in achieving positional constraints. The authors provide simulation results that illustrate the efficacy of their results.

The paper, \cite{7113913}, deals with the control of an n-link robotic manipulator with input saturation using adaptive impedance neural network control. The neural network model, employed in this approach, uses radial basis functions. The adaptive neural impedance controllers are designed using Lyapunov's method. Both full state and output feedback controllers are considered in this work along with an auxiliary signal specially designed to handle the saturation effect of a robotic manipulator.

The paper, \cite{Xu2015}, presents the adaptive dynamic surface control of a hypersonic flight vehicle in an environment with unknown dynamics and input non-linearities. In this framework, the authors utilize a radial basis function based neural network control regime to estimate the control gain function and avoid the singularity problem. The authors also make use of dynamic surface control and the Nussbaum function  in order to construct the controller.  Their approach relies on functional decomposition where a PID controller is utilized for the velocity subsystem and the neural network control regime deals with the attitude subsystem. In this regime, stability is guaranteed via the Lyapunov approach and simulation results demonstrate the efficacy of their approach.

In \cite{HE20141843}, Wei He et al. present the  full-state and output feedback control of a thruster-assisted single-point mooring system in the presence of uncertainties and unknown backlash-like hysteresis nonlinearities. Their regime is based on the backstepping technique, Lyapunov synthesis, and adaptive neural network  control. The backlash-like hysteresis is controlled using backstepping by modeling the hysteresis as a linear term and a bounded non-linear term. The neural networks are used to estimate unmeasurable states. Simulation results and further analysis demonstrate the feasibility of this approach. Moreover, the authors prove that their control schemes are semi-globally uniformly bounded.

The paper, \cite{He2015}, presents a neural network based control regime for a robot with unknown system dynamics. The neural networks are used to deal with uncertainties in the system environment and to approximate the unknown model of the robot. Both full-state and output feedback control are utilized in this work. The authors demonstrate uniform ultimate boundedness using Lyapunov’s approach and further demonstrate convergence towards a small neighborhood of zero with an appropriate choice of design parameters. Simulation results demonstrate the effectiveness of their approach.


\subsection{Stability Assurance while Learning the Optimal Policy}

In optimal control theory, a cost function is given as $J(x,u) = \int_0^T r(x(t),u(t)) dt$ and an optimal controller is sought that minimizes $J$ while adhering to the dynamic constraints $\dot x = f(x) + g(x)u$ with $x(0) = x_0$ \cite{liberzon2011calculus}. When $T = \infty$ and the dynamic constraints are autonomous, the optimal value function $V^*(x_0) := \inf_{u \in \mathcal{U}} J(x(\cdot;x_0), u)$ is time invariant and satisfies the Hamilton-Jacobi-Bellman (HJB) equation \[ 0 = r(x,u^*) + (\Delta_x V^*)^T(f(x) + g(x)u^*), \] where $u^*$ is the optimal controller \cite{liberzon2011calculus,kamalapurkar2018reinforcement}. When $r(x(t),u(t)) = x^T(t) Q x(t) + u^T(t) R u(t)$, the optimal controller can be expressed, in terms of the optimal value function, as \[u^*(x) = -\frac12 R^{-1}g^T(x)\Delta_x V^*(x).\]

However, there is a limited collection of dynamics for which a closed form solution to the HJB is known \cite{lewis2012optimal,liberzon2011calculus}. This motivates the investigation into numerical methods for determining the value function and subsequently an optimal controller. For offline optimal control problems, value and policy iteration \cite{bertsekas2005dynamic} are valuable tools for determining an approximation of the value function and optimal controller.

However, it is desirable to learn the optimal value function and optimal controller online for many situations, such as when there are uncertainties in the dynamics that cannot be accounted for in an offline setting or in the presence exogenous disturbances \cite{vrabie2009neural}. There are several extant methods that learn the optimal controller online through the, so called, method of adaptive dynamic programming (ADP) that use an online actor-critic architecture for learning the optimal value function \cite{vamvoudakis2010online,stingu2011approximate,kamalapurkar2018reinforcement}. For example, the work of \cite{stingu2011approximate} uses a radial basis function network for approximating the value function, while \cite{kamalapurkar2015approximate} uses a polynomial basis. A local learning method was introduced in \cite{kamalapurkar2018reinforcement,kamalapurkar2016efficient} to reduce the computational overhead required for an accurate approximation.

Online learning control problems face the issue of stability during the transient learning phase \cite{lewis2012optimal,kamalapurkar2018reinforcement}. As the weights of a neural network are tuned to improve the accuracy of the value function approximation and thus the accuracy of the controller, uncertainties can lead to unbounded trajectories and a loss of stability. Two main approaches have been utilized in the literature to provide assurances of stability during the learning phase of the optimal control problem. One approach uses the concept of persistence of excitation \cite{vamvoudakis2010online}, while the other utilizes concurrent learning \cite{kamalapurkar2016efficient,Kamalapurkar2013ConcurrentLA}. The implementation of both are used in tandem with Lyapunov stability analyses to provide stability guarantees \cite{vamvoudakis2010online,kamalapurkar2016efficient}.

%% file: learning.tex
\section{Learning: Specification Inference and Learning}

In \cite{JhaTeLEx2017}, Jha, Tiwari, Seshia, Sahai, and Shankar present an approach to passive learning of Signal Temporal Logic (STL) formulas that only observes the positive examples and excludes systemic safe active experimentation, which is outside of the traditional two categorizations. In practice, there are classifier-learning and active-learning techniques. Classifier-learning techniques use both positive and negative examples for STL formula learning and active-learning techniques experiment on the system to extract counterexamples. The authors evaluate their system on a metric of tightness, which represents predicates and temporal operators as smooth functions and is influenced by techniques for numerical optimization based on gradients. Further work would be parameter specification for tightening, examine proposed additional metrics, and parallelize the computation of the trace metrics. 

In \cite{JhaBoolean2017}, Jha, Raman, Pinto, Sahai, and Francis present an approach to Boolean formulae learning that queries example-labeling oracles. The main contribution of the paper is an algorithm for Boolean formulae learning with a given level of confidence. The aforementioned algorithm can be used to generate explanations for decisions made by Artificial Intelligence (AI) based systems. The benefits of making the self-explanation capabilities for AI systems would be easing the human-system interaction. Further work should include the utilization of a language with richer logic for the explanations involving real numbers and the generation of multiple valid justifications for any given inquiry. 

In \cite{FisacSafety2017}, Fisac et al. present a Hamilton-Jacobi reachability methods-based framework for general safety that works with any learning algorithm. The Gaussian processes and the reachability analysis used in the framework are computationally expensive and are not scalable. The main control theory aspect utilized in the framework is robust reachability combined with observational Bayesian analysis. The authors provide proof that the two methods in conjunction work well. Further work should include increasing the safety certificates with respect to intelligent systems.

In \cite{HerbertFaSTrack2017}, Herbert et al. present an algorithm for fast and safe tracking (FaSTrack) in high dimensional systems called FaSTrackHD. The FaSTrack framework can be used in combination with a path planner to guarantee the safety and efficiency of the plan. This paper provides a look-up table tool, a framework for tool implementation, and a high-dimensional demonstration of the tool and framework combination. FaSTrack makes planners more robust while maintaining fast computation. Further work could include moving obstacle robustness, considering environmental disturbances to which the error bounds should adapt, and multi-planner demonstrations.

In \cite{AkametaluReachability2014}, Akametalu et al. present an approach that utilizes a Gaussian process model and maximal safe set approximations to learn a system's unknown characteristics. Safety is incorporated in the metric through which the learning performance is evaluated, thus reducing controller switching. The safe learning algorithm presented is reachability-based. Through utilizing Gaussian processes, it learns a model of environmental disturbances. Overall, the algorithm provides better safety guarantees than the state-of-the-art frameworks. Further work could include safe set real-time updates and using sparse Gaussian processes to speed up the computations.

In \cite{FisacPragmatic2017}, Fisac et al. present a solution to the value-alignment problem based on learning models that can be experientially verified. The framework interweaves human decision making and cooperative inverse reinforcement learning to solve the problem of value-alignment. This work shows that cooperative inverse reinforcement learning problems can be solved not just theoretically but practically. Further work should include making the approaches more efficient and utilizing human models that are more realistic. 

In \cite{ChenUAVSafety2017}, Chen et al. present an approach of creating platoons of unmanned aerial vehicles (UAVs) that uses Hamilton-Jacobi reachability analysis. In order to implement a sense of structure in the air, the approach requires that the platoons of UAVs fly on air highways. The authors verify their concept through simulating forming a platoon, an intruder vehicle, and transitioning between highways. The fast-marching algorithm is used to determine the air highway placements. The implemented algorithm takes an arbitrary cost map and determines the path set from the origin to the destination. As long as only one breach per platoon vehicle occurs, the controller assures that collisions will not occur with one altitude level. The addition of safety breaches requires additional altitude levels.  

In \cite{ChenSafePlatooning2017}, Chen, Hu, Mackin, Fisac, and Tomlin present an approach of safe and reliable platoons of unmanned aerial vehicles (UAVs) that uses Hamilton-Jacobi reachability analysis which is the initial work or a simplification of \cite{ChenUAVSafety2017}. In order to implement a type of structure in the air, the approach requires that the platoons of UAVs fly on air highways. They verify the concept through simulating a platoon formation, a vehicle intrusion, and transitioning between highways. The fast-marching algorithm is used to determine the air highway placements by taking an arbitrary cost map and determining the path set from the origin to the destination. As long as only one breach per platoon vehicle occurs, the controller assures that collisions will not occur with one altitude level. The addition of safety breaches requires additional altitude levels.  

In \cite{jin2015mining}, Jin et al. present a scalable framework to mine specifications from a closed-loop control model of an industrial system based on the system's behavior. The input is represented by a parametric signal temporal logic (PSTL)\cite{asarin2011parametric}, where time or scale values are unknown parameters. Based on the quantitative semantics of signal temporal logic (STL), a falsification engine in \cite{donze2010breach} is applied to check for the existence of traces violating current STL formulae. To find the strongest candidate specification, a parameter synthesis algorithm based on $\delta$-tight valuations is applied by finding the valuation closest to the boundary of a validity domain. Since the monotonicity of a PSTL formula will greatly decrease the computational complexity, they also propose an algorithm to check and take advantage of it. The framework's utility and scalability are tested on an automatic transmission model, an air-fuel ratio control model, and a diesel engine model.

\subsection{Automata Learning}
Automata Learning can be partitioned into active and passive learning approaches. Automata can be discrete, continuous, or hybrid in nature. Discrete automata are state machines where the states are discrete, whereas continuous automata are state machines where the states are continuous. Hybrid automata are state machines where the variables are set by ordinary differential equations. Active Learning allows the learner machine to decide from which data to learn \cite{SettlesActiveLearning2010} \cite{OlssonActive2009}. Active Learning is utilized in synthesizing membership queries, sampling based on selective streams, and sampling that is based on a data pool. 

The remainder of the section is split into learning hybrid automata, timed automata learning, discrete automata learning, and active learning of automata.

In \cite{AngluinQuery1988}, Angluin studies and describes multiple types of queries in formal domains. They examined membership, equivalence, subset, superset, disjointness, and exhaustiveness queries. They applied the aforementioned types of queries to various domains. Singleton languages, \textit{k}-CNF formulas, \textit{k}-DNF formulas, monotone DNF formulas, regular languages, \textit{k}-bounded CFLs, \textit{k}-term DNF formulas, \textit{k}-clause CNF formulas, pattern languages, very restricted CFLs, and double sunflower are the domains examined in the paper. Their results show that some queries perform better on certain domains. 

In \cite{BalkanUnderminer2017}, Balkan, Tabuada, Deshmukh, Jin, and Kapinski present the framework, Underminer,  for determining which behaviors are non-convergent in the designs of control systems that are embedded. The system being examined by the framework is considered a black-box, meaning that only the inputs and outputs are examined. The framework uses Convergence Classifier Functions (CCFs) to distinguish between the non-convergent behaviors and convergent behaviors. The framework can be applied to a system either in the early development stages or late in the controller development stage. The Underminer framework first determines the non-convergent behaviors using CCFs and then performs CCF automated test generation and examination. Temporal logic, Lyapunov functions, and machine learning classification techniques are all CCFs supported by Underminer. Optimizer-driven testing and sampling-based test generation techniques are utilized by Underminer during the test generation phase. 

\subsubsection{Hybrid Cases}

In \cite{AnsinAutomated2013}, Ansin and Lundberg present the hybrid congruence generator extension (HyCGE) learning algorithm and examine its characteristics. The validity and accuracy of the algorithm are also graphically demonstrated. The ratio of the completeness was shown to have continuous growth for the example automata. A future application of the HyCGE algorithm is physical systems learning. Further work could include full diagrams and simulations of the human body through learning that is automated. The HyCGE learning algorithm can be used for the aforementioned automated learning.

In \cite{MedhatFramework2015}, Medhat, Ramesh, Bonakdarpour, and Fischmeister present an approach for hybrid automata learning based on the implementations of control systems as black-boxes and examine the inputs and outputs of said systems. Input/output events are gathered from the observed traces, based on which target system is modeled with a Mealy machine. The dynamics of state variables in each discrete state are analyzed, where the time factor is involved. The framework is limited in that all of the types of signals may not be supported, hidden states cannot be observed, guards using output values for transitions can not be modeled, and that change from input to output is assumed instantaneous. Further work could include improving data preprocessing and segmentation; utilizing algorithms that have better efficiency; support for causality rules expansion; and automata inference improvements. 

In \cite{NiggemannLearning2012}, Niggemann, Stein, Maier, Voden\v{c}arevi\'{c}, and B\"{u}ning present and analyze the Hybrid Bottom-Up Timing Learning Algorithm (HyBUTLA), an algorithm for automata learning. The observed traces are segmented according to abrupt change in state values. The dynamics of each segments is modeled using linear regression or neural networks and is represented as a function within a state. Similarity of two states is tested by checking the probability for staying in the state versus a specific transition occurring. In a new bottom-up merging order, similar states get merged and the higher efficiency in comparison to traditional top-down order is verified. The scalability of HyBUTLA was tested using generated artificial data. Further work could include developing a better algorithm for hybrid state identification, using probability density functions instead of fixed time intervals, and further algorithmic plant applications.

In \cite{SummervilleCHARDA2017}, Summerville, Osborn, and Mateas exhibit and evaluate the Causal Hybrid Automata Recovery via Dynamic Analysis (CHARDA) framework. The framework performs modal identification and learns the causal guards. In the first phase, one set of potential linear model templates has a cost function with penalty criteria applied to segment traces by switch-points and select the optimal models. For the guard condition, they applied Normalized Pointwise Mutual Information (NPMI) to select predicates from a predefined set for each mode transition. They tested the CHARDA framework on Super Mario Bros as a novel domain. Further work, with respect to the segmentation, could include testing the segmentation and then transition learning general framework to other techniques, determine the optimal method of differentiating between similar modes, and incorporate other approaches' techniques. Further work, for learning causal guards, could include testing it on other domains and improving upon the analytic precision. 

In \cite{BogomolovPDDL+2015}, Bogomolov, Magazzeni, Minopoli, and Wehrle examine the theoretical basis for translation using Planning Domain Definition Language+ (PDDL+). The translation of must-transition automata to only may-transition automata is the prime directive of their schema. PDDL+ is used since processes and events are well-handled. The construction of the translated hybrid automata causes the state space that is reachable to be the same as hybrid automata that are linear. Hybrid automata with dynamics that are affine have a reachable state result that is over-approximated. This paper founds the process of PDDL+ problem translation into hybrid automata that are standard. 

In \cite{grosu2007learning}, Grosu et al. propose a methodology for inferring cycle-linear hybrid automata for two types of excitable cells from virtual measurements generated by existing nonlinear models. In the first phase, traces of electrical signal, known as action potential with respect to time, are segmented by filtered null points and inflection points. Then, they apply modified Prony’s method to fit an exponential function to each of the segments within a predefined error bound, which derives a flow condition for each mode. The transition voltages for mode switches are approximated by a range of voltage obtained from transitions’ post states. In the second phase, all of the inferred linear hybrid automata are merged into a signal one by setting an epoch transition. The accuracy of the proposed method is tested by comparing it with the simulations of two existing models.

\subsubsection{Timed Automata}

In \cite{IegorovPeriodic2017}, Iegorov, Torres, and Fischmeister present PeTaMi (PEriodic TAsk MIner), a periodic task mining approach and tool that utilizes the information from the real-time systems being examined. The tool classifies the tasks into periodic and non-periodic and then, using only the periodic tasks, determines the time of response and periods for said tasks. This method is automatic and uses the timestamped traces of events for classification. The period and time of response for a periodic task is determine through the use of a clustering-based mining approach for temporal specifications. The tool examines the input trace of execution and trace fields and then classifies them into periodic or non-periodic. In order to evaluate the system, they tested it on an unmanned aerial vehicle (UAV) case study and a commercial car in operation case study. Those results were indicative of information needed to determine the period and response time for the periodic tasks. 

In \cite{VerwerEfficient2010}, Verwer presents an extensive study of the timed automata identification complexity theory. They also unveil an algorithm that uses labeled data to perform timed automata identification. The algorithm is then tested on real-time system identification with a behavior system for truck drivers. It was determined that deterministic timed automata (DTA), each with a single clock, have efficient identification, whereas DTA with more than one clock are not efficiently identifiable. They discuss the Real-Time Identification (RTI) algorithm which is efficiently used for deterministic real-time automata identification using only a input sample that is timed and labeled data. It was tested on artificial data. The RTI algorithm was tested on the truck driver system and had 80 percent accuracy. Further theoretical work could be focused on the theory of identifying timed automaton, algorithms for identifying timed automaton, or evaluating probabilistic models. Some further applications include behavior identification, visualization providing insight, or  model checking identification.  

In \cite{FiterauCombining2016}, Fiter\u{a}u-Bro\c{s}tean, Janssen, and Vaandrager use a combination of model learning and model checking to infer models and thoroughly examine said models. They studied the Windows, Linux, and FreeBSD TCP (Free Berkeley Software Distribution Transmission Control Protocol) implementations. The TCP learning implementation includes upper layer calls, data transmission, servers and inferred TCP models, and learned models for all of the aforementioned operating systems. Model checking is used to fully examine the explicit models of the components. The described method is versatile and has variable applications. Further work could include making the construction of TCP abstractions completely automatic and creating state machine generation capable learning algorithms. 

Verwer, de Weerdt, and Witteveen present the efficiency of timed automata learning \cite{VerwerEfficiency2011}. They show that deterministic timed automata (DTAs) in general cannot be efficiently learned using labeled data but that one-clock DTAs (1-DTAs) can be efficiently learned using labeled data. They establish an algorithm for learning one-clock DTAs. Further work could include using an n-DTA identification algorithm to efficiently identify 1-DTAs and determining the largest class of efficiently learnable DTAs. 

In \cite{VerwerEfficiently2012}, Verwer, de Weerdt, and Witteveen present a deterministic real-time automaton (DRTA) learning algorithm, real-time identification (RTI) algorithm, utilizing event sequences that were labeled and time-stamped. The algorithm merges states based on the evidence to identify a deterministic finite-state automaton (DFA). The performance of the aforementioned algorithm is evaluated using artificially generated data. It is evident that it is harder to identify a DFA than to identify an equivalent DRTA. Further work could include extending the RTI algorithm to work for unlabeled data, enable identification of timed automata that are more general, and evaluate the performance of the adapted algorithms. 

In \cite{ZhangCar2017}, Zhang, Lin, Wang, and Verwer present a car-learning behavioral model that utilizes automata learning algorithms. Driving patterns are discovered from examining and extracting common sequences of states from the model. The model was trained and tested on the Next Generation SIMulation dataset exhibiting high accuracy. They also represent time data that is multivariate as strings that are symbolic and timed. The input data clustering utilizes the properties of the temporal processes. The mentioned utilizations significantly increased the accuracy of the data fitting. Further work could include comparing the model to other existing model implementations, apply to drivers' decision making, and assisting in the development of a car-following controller.  

In \cite{AartsLearning2012}, Aarts, Kuppens, Tretmans, Vaandrager, and Verwer demonstrate that the combination of tools for verification and testing software can be used in combination with active learning in an industrial environment. The tool combinations can be utilized to evaluate and improve upon the active learning results. They employ active state machine learning to perform conformance tests on implementations of the protocol for bounded retransmission. Further work could include improving upon the Markov Reward Model Checker (MRMC), TorXakis, and Tomte tools. 

In \cite{SmeenkApplying2015}, Smeenk, Moerman, Vaandrager, and Jansen present control software from industry with automata learning actively applied. They also examined LearnLib's capabilities with regards to learning the model of the Engine Status Manager (ESM). Due to the large number of abstractions required by LearnLib, the authors extended a preexisting algorithm to enable the computation of a sequence that is adaptive distinguishing. Using the algorithm extension, they learned a model using a Mealy machine for the ESM. The extended algorithm performed superiorly to other algorithms for conformance testing. Further work could include model learning for multi-functional ESM and model learning for ESM connecting components.  

In \cite{VerwerAlgorithm2007}, Verwer, de Weerdt, and Witteveen exhibit a real-time automata learning algorithm. The traditional algorithms involve state merges but their algorithm implements state splits, which utilize the time data. Real-time automata are prevalent in models of systems where behavior monitoring is vital. Further work could include extending the algorithm to cover probabilistic timed automata which would only require examining the positive examples and enable real-world applications. 

\subsubsection{Discrete Learning}

In \cite{MuddassarIDS2012}, Sindhu and Meinke present the incremental distinguishing sequences (IDS) algorithm for learning deterministic finite automata (DFAs) incrementally using sequence differentiation. The performance analysis looked mainly at the variety of learning queries possible and the time taken for the learning. From the analysis, the algorithm is most applicable for problems in software engineering.

In \cite{GiantamidisLearning2016}, Giantamidis and Tripakis present three algorithms for solving the problem of Moore Machine learning from input-output traces and formalize said problem. The first algorithm is the prefix tree acceptor product (PTAP) algorithm which, using self-loops, completes the Moore Machine formed from input-output trace data. The second algorithm is the PRPNI algorithm that utilizes the Regular Positive and Negative Inference (RPNI) algorithm for learning an automata product and encodes the Moore Machine. The third algorithm mentioned is the MooreMI algorithm which performs Moore Machine learning directly utilizing the state merging from the extended PTAP algorithm. Overall, the MooreMI algorithm is shown to be the most progressive algorithm of the set. Further work could include learning other types of state machines, making each algorithm incremental, more experimentation and implementation, and extending the algorithms to model any black-box system.

\subsubsection{Active Learning}
In \cite{RaffeltLearnLib2005}, Raffelt, Steffen, and Berg present LearnLib, a library for learning and experimenting on automata. The library has a modular structure conducive to scenario tailoring. Users have the ability to specify the model structure, the applied techniques for optimization, choose between the automatic and the interactive mode, and use a simulator for random models. The library examines the power of exploiting optimization properties. Further work could include extending the library by modular increases, other property exploitation, exploring other occurrences, and increasing the complexity of the scenarios of applications. Works also will include making LearnLib available online. 

In \cite{MertenNextGeneration2011}, Merten, Steffen, Howar, and Margaria present Next Generation LearnLib (NGLL) framework. The framework is intended to make the task of setup creation for fitting learning simpler. It allows for remote component integration into the framework so it can be used globally. NGLL has tools and methods available to perform a multitude of tasks, such as harnesses for testing a system, techniques for making a system more abstract or refining said system, and mechanisms for resetting a system. The framework was designed for ease of use with respect to controlling, adapting, or evaluating a system. 

In \cite{IsbernerOpenSource2015}, Isberner, Howar, and Steffen exhibit the new open-source version of LearnLib, an active automata learning library. It allows for visualization learning algorithm in-detail progress. The tool has a modular design which simplifies modifying features of the algorithm. The feature component, AutomataLib, is a finite-state machine data structure and algorithmic toolkit. The performance of the new LearnLib is efficient while allowing abstractions on a high-level. Further work could include adapting LearnLib Studio for the new LearnLib and extending the capabilities to enable register automata learning.

\subsection{Safe Reinforcement Learning}
Reinforcement Learning (RL) is a widely adopted frameworks in Artificial Intelligence for producing intelligent behavior in autonomous agents \cite{KaelblingSurvey1996}. 
\noindent By interacting with their environments, these autonomous agents can learn to execute optimal behaviors via trial and error over a specified time period\cite{Arulkumaran2017}. Thus, in this scheme, agents can be programmed to accomplish a wide range of tasks without forcing the programmer to delineate how these tasks should be carried out. Furthermore,  reinforcement learning has exhibited great success in modeling interactions between an autonomous agent and its environment where a model is too difficult to procure. In fact, reinforcement learning has demonstrated great utility in contexts such as game playing, robotics applications, industrial automation, autonomous vehicles, complex financial models, and structured prediction \cite{KaelblingSurvey1996,BerkenkampSafeModelRL2017,Arulkumaran2017}.
\par Although there has been great success in using reinforcement learning to solve low dimensional problems, the majority of learning approaches exhibit scalability issues. However, with the advent of successful deep learning paradigms there has been great research impetus in improving the scalability of reinforcement learning schemes through the use of neural networks. Using this framework, the field of Deep Reinforcement Learning has displayed great promise in accelerating progress within the reinforcement learning community by adequately addressing the prohibitive dimensionality issues associated with complex state action spaces \cite{Arulkumaran2017}. Coupled with the desire to enhance the efficiency of reinforcement learning regimes is the desire to guarantee safety, particularly in safety critical environments where there is significant risk towards the people and environments involved \cite{BerkenkampSafeModelRL2017}. As with many algorithms in machine learning, the majority of traditional reinforcement learning techniques are unable to provide guarantees about the safe operation of a system. However, in order to deploy this kind of software in safety critical settings, one must be able to provide formal arguments about the safe operation. Thus, in recent years there has been a great deal of research into developing safe reinforcement learning frameworks. 
\par In a classic reinforcement-learning architecture, an agent interacts with its environment over a given time period through observations and actions that are typically described by a Markov Decision Process (MDP) \cite{Arulkumaran2017}. During each time step an agent receives some information about the current state of its environment and chooses an action to perform generating an output about its altered state. Based on each action selection, a reinforcement signal denoting the value of this transition is communicated to the agent. Throughout the learning process, the agent is trained to select actions that maximize the long-term sum of reinforcement signals \cite{KaelblingSurvey1996}. There are numerous algorithms that allow an agent to select an optimal behavior policy through systematic trial and error and a survey of these techniques can be found in the following papers \cite{KaelblingSurvey1996,Arulkumaran2017}. Formally, the Markov Decision Process model can be defined as a tuple (\textit{S,A,T,R, $p_0$}) where:
\begin{itemize}
\item \textit{S} is a continuous set of states,
\item \textit{A} is a set of continuous actions, 
\item \textit{T}: $S \times A \rightarrow S$ is a probabilistic transition function that maps a state action pair at some time $t$ onto a probabilistic distribution of states,
\item \textit{R}: $S \times A \times S \rightarrow \mathbb{R}$ is an immediate reward function,
\item and $p_0$ specifies a unique initial state \cite{AlshiekhSafeRL2017}.
\end{itemize}
Thus, the problem of reinforcement learning is to learn a policy $\pi_\theta: S \times A \rightarrow \mathbb{R}$ that mamximizes the cumulative discounted reward $\sum_{t=0}^{T-1} \gamma^TR(s_t,r_t,s_{t+1})
$ where \textit{R} is the reward at time step $t$ and $\gamma \in [0,1]$ is a discount factor that controls the influence of future rewards \cite{PintoRobust, QingkaiAcceleratedSafeRL2018}. Additionally, in most frameworks, there is a common assumption that the learning environment is non-deterministic in nature. Thus, the same choice of action by an agent may result in different reward values at different time instants \cite{KaelblingSurvey1996}. However, this assumption is often coupled with the presumption that the probabilities of making state transitions or observing certain reinforcement signals do not change over time \cite{KaelblingSurvey1996}. Despite the success of the MDP reinforcement learning model, this approach has several limitations. The MDP model suffers from representation limitations and not all learning agents can estimate a notion of a Markovian state \cite{ShalevMultiRL}. This shortcoming in representation is commonly found in robotics applications. Additionally, in some applications such as autonomous driving, the transition of states is dependent not only the agents actions but also on the actions of other agents. Moreover, in these cases, we must be able to guarantee safety during the exploration process in order to prevent dangerous and costly adverse scenarios. In response to the modeling, safety, and performance challenges present within the reinforcement learning community, numerous researchers have proposed various solutions aimed at generating effective, robust, and intelligent behavior  \cite{ShalevMultiRL}.
\par In a 2015 survey of existing safe reinforcement learning literature, Javier Garc\'ia defined safe reinforcement learning as "the process of learning policies that maximize the expectation of the return in problems in which it is important to ensure reasonable system performance and/or respect safety constraints during the learning and/or deployment processes \cite{GarciaSurvey}." Moreover, in his work, Garc\'ia notes that the idea of risk in safe reinforcement learning regimes is complex and takes many forms. In many works, risk is related to the presence of uncertainties in the environment with which an autonomous agent interacts. Thus, even if an agent is able to correctly learn an optimal behavior policy there are still risks associated with unexpected disturbances that may result in generalization and simulation-transfer issues \cite{GarciaSurvey}. Additionally, one must reason about the failure modes of learned policies in order to identify behaviors that may pose considerable danger to the agents and environments considered. In his survey, Garc\'ia identifies two fundamental approaches towards the problem of creating safe reinforcement learning architectures present in the research literature. The first set of methods alter the optimization criteria present in the learning process by incorporating a notion of risk into the reward optimization process. The second set of procedures modifies the exploration process of the learning agent either by including the likelihood of entering error states through the use of external knowledge or via the use of a risk metric \cite{GarciaSurvey}. Garc\'ia's survey provides a great review of these works and discusses several novel directions for researchers interested in the field. 
\par As mentioned above, one of the central foci in safe reinforcement learning algorithms is ensuring that agents do not behave in unexpected and errant ways during the exploration and learning processes. In many cases, the exploration process may be costly or dangerous and we are required to ensure safety at all times. Addressing this issue in their work, Theodore Perkins et al. propose a method, in \cite{PerkinsLyapunov}, for constructing safe and reliable learning agents for control problems using Lyapunov design principles. In their framework, an agent learns an optimal behavior by switching between a set of base level controllers designed using Lyapunov domain knowledge. Thus, the authors are able to ensure safe control by limiting the set of actions that an agent can select from in order to guarantee that every transition descends towards a Lyapunov function \cite{PerkinsLyapunov}. In fact, in the experimental evaluation of their techniques, the authors were able to demonstrate that this methodology was feasible regardless of the particular reinforcement learning algorithm selected. In a similar work introduced by Felix Berkenkamp et al. \cite{BerkenkampSafeModelRL2017}, the authors extend the results produced in the field of Lyapunov stability verification in order to obtain control policies that have provable stability and safety guarantees\footnote{An implementation of Felix Bergenkamp's methodology can be found at \url{https://github.com/befelix/safe_learning}}. In fact, the authors demonstrate that it is possible to learn an optimal policy with stability and safety guarantees while striving not to limit the state space from which an agent may select an action. Moreover, their approach allows an agent to collect data in order to expand the state space of the computed safe region while also improving the control policy \cite{BerkenkampSafeModelRL2017}. This is done by starting from a configuration that is assumed to be stable and gathering data at safe and informative points in order to improve the control policy using the newly obtained, enhanced model of the system. 
\par While the above approaches display promising results for achieving safe control, there are concerns that methods that limit the environment's state space may eliminate actions that are better in achieving an optimal cost. Additionally these methods require more data than traditional reinforcement learning regimes. Thus, these models may be more susceptible to model inaccuracies due to insufficient data and model disturbances \cite{GarciaSurvey,ThomasSafeRL,MunosSafeEfficientRL2016}. In support of these concerns, in their paper, Anayo Akametalu et al. cite that in reachability based algorithms, the computed safe region utilized in safe reinforcement learning algorithms may not accurately capture the true nature of disturbances present in an environment and therefore lead to safety guarantees that are no longer dependable. Thus, to address these concerns, the authors propose a learning regime that makes use of a Gaussian process to learn a system's unknown dynamics by iteratively approximating a safe set. Throughout the learning process, their safety model is validated online in real time allowing the authors to guarantee safety in conditions where the assumptions about environmental disturbances are incorrect. Moreover, their methodology makes use of a safety metric in order to limit the amount of time that their system interferes with the learning process. Their framework represents an interesting approach that proposes a reachability analysis scheme that cross validates the model online. In a similar work by Jeremy H. Gilula et al. \cite{GillulaGuaranteedSafeRL2012}, the authors present a guaranteed safe online learning framework for an aerial robot with a limited field of view. Their approach combines reachability analysis and online reinforcement learning. Simulation of their methods demonstrates that combining the two paradigms allows for high performance and safety.
\par Another promising area of safe reinforcement learning makes use of model checking algorithms and methods derived form formal verification to ensure safety. In their paper, Mohammed Alshiekh et al. present a temporal logic based approach for learning optimal policies \cite{AlshiekhSafeRL2017}. To ensure that a system specification expressed in temporal logic is satisfied, the authors implemented a system, called shield, that alters the learning process in two fundamentally different ways. In the first regime, shield provides a list of safe actions from which an agent may select. In the second regime, shield monitors the actions of the learning agent and interferes only when necessary \cite{AlshiekhSafeRL2017}. In another work presented by Shashank Pathak et al. \cite{PathakSafeRLVerification2018}, the authors make use of probabilistic model checking algorithms in order to verify and repair learned policies. In this work, the authors provide a set of repair approaches and policy modification strategies that ensure that the probability of reaching unwanted states and executions, once the learning process terminates, remains within an acceptable bound. Additionally, the authors provide a benchmark, called "Slippery Grid World," that other authors may use to assess the scalability and effectiveness of their approaches \cite{PathakSafeRLVerification2018}. In a similar work done by Nathan Fulton and Andr\'e Platzer, the authors propose a technique, called Justified Speculative Control (JSC), for provably safe reinforcement learning \cite{FultonSafe18}. Their results combine formal verification and runtime monitoring in order to ensure the safety of a system. Moreover, their work contributes a set of proofs that transfer computer-checked safety proofs for hybrid dynamical systems to policies obtained by generic reinforcement learning algorithms \cite{FultonSafe18}. In their learning environment, verification results are maintained when learning agents limit their operation to remain within verified control choices. Thus, their regime is similar to many works cited in Garc\'ia's survey in which incorporating knowledge about safe control into the learning system preserves safety guarantees \cite{FultonSafe18,GarciaSurvey}.
\par In recent years, researchers have demonstrated that machine learning models may be susceptible to attacks during all phases of the learning process \cite{KumarAdversarial2017}. The majority of models can be fooled by carefully crafted input samples that result in misclassification. Using these samples, an attacker may be able to slow the learning process or affect the performance of a model by causing the system to enter into a dangerous configuration. These concerns have served as the motivation for an emerging field of research called Adversarial Machine Learning \cite{KumarAdversarial2017}. While adversarial machine learning approaches usually deal with perception tasks in neural networks, there have been a few works proposed within the safe reinforcement learning community: particularly tasks related to autonomous vehicles and neural network aided reinforcement learning. In a paper proposed by Lerrel Pinot et al., the authors present a scheme for robust adversarial reinforcement learning \cite{PintoRobust}. They argue that numerous reinforcement learning regimes suffer from a high reliance on data causing them to suffer from generalization issues and errant behavior caused by the presence of environment disturbances. Thus, in order to design more disturbance-robust systems, the authors train agents in the presence of an adversarial agent that delivers model disturbances aimed at destabilizing the system \cite{PintoRobust}. Using a zero sum minmax objective function to learn a policy, the authors were able to successfully train an agent that was robust to the adversarial agent as well as other environmental disturbances. Moreover, their agent was more robust to differences in the testing and training environments. In a similar work done by Henrik Aslund et al., the authors present the concept of virtuously safe reinforcement learning which is defined as resilience toward perturbed perception, safe exploration, and safe interruption \cite{AslundVirtuouslySafeRL2018}. In their work, the authors demonstrate how to create a learning scheme in which agents are able to manage adversaries using four exploration strategies that accomplish the goal of virtuously safe reinforcement learning. 
\par In summary, Safe Reinforcement Learning techniques are used to address a wide range of problems where it is important to guarantee safety constraints of learning policies. There are numerous approaches with various drawbacks, advantages, and different conceptions of risk that have demonstrated great promise in ensuring that an agent will operate safely within its environment. However, there are still many open questions within this realm, such as balancing safety constraints with performance desires and improving the scalability and computational tractability of the current approaches 

%% file: verification.tex
\section{Verification for AI/ML Components and Systems}

\subsection{Verification: Neural Network Reachability}

Artificial neural networks are used in systems that introduce machine learning components to resolve complex problems. This can be attributed to the impressive ability of neural networks to approximate complex functions as shown by the Universal Approximation Theorem \cite{hornik1989multilayer}. Neural networks are trained over a finite amount of input and output data, and are expected to generalize said data and produce desirable outputs for the given inputs and even for previously unseen inputs. The data-driven nature and lack of efficient methods for analysis of neural networks leads to, in most cases, the treatment of neural networks as black boxes with no assurance in safety. However, due to the rapid development artificial intelligence inspired applications, neural networks have recently been deployed in  several safety-critical systems such as self-driving vehicles \cite{bojarski2016end}, autonomous systems \cite{julian2017neural}, and aircraft collision avoidance procedures \cite{julian2016policy}. Regrettably, it has been observed that neural networks can react in unexpected and incorrect ways to even slight perturbations of their inputs \cite{szegedy2013intriguing}. Therefore, there is an urgent need for methods that can provide formal guarantees about the behavioral properties and specifications of neural networks, especially for the purpose of safety assurance \cite{Leofante2018}.

Verifying neural networks is a hard problem, and  it has been demonstrated that validating even simple properties about their behavior is an NP-complete problem \cite{katz2017reluplex}. The
difficulties encountered in verification mainly arise from the presence of activation functions and the complex structure of neural networks. Moreover, neural networks are large-scale, nonlinear, non-convex, and often incomprehensible to humans. The action of a neuron depends on its activation function described as $y_i = f(\sum\nolimits_{j=1}^{n}\omega_{ij}x_j+\theta_i)$, where $x_j$ is the $j$th input of the $i$th neuron, $\omega_{ij}$ is the weight from the $j$th input to the $i$th neuron, $\theta_i$ is called the bias of the $i$th neuron, $y_i$ is the output of the $i$th neuron,  and $f(\cdot)$ is the activation function. Typically, the activation function is either the rectified linear unit, logistic sigmoid, hyperbolic tangent, the exponential linear unit, or another linear function. 

\begin{figure}
	\begin{center}
	\includegraphics[width=9cm]{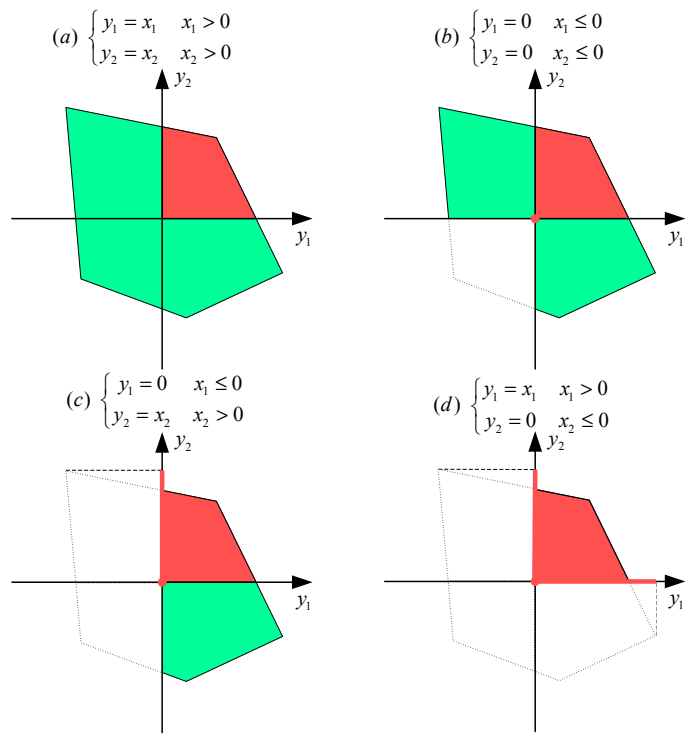}
		\caption{A visualization of the operations on polytopes using ReLU functions in \cite{xiang2017reachable_arxiv}.}
		\label{fig:ReLU}
	\end{center}
\end{figure}

To circumvent the difficulties brought by the nonlinearities present in the neural networks, the majority of recent results focus on activation functions of piecewise linear forms, $f(x) = \max(0,x)$, and in particular the Rectified Linear Unit (ReLU). For instance in \cite{xiang2017reachable_arxiv}, by taking advantage of the piecewise linear nature of ReLU activation functions, the output set computation can be formulated as operations of polytopes if the input set is given in the form of unions of polytopes. Figure \ref{fig:ReLU} shows a visualization of the polytopic operations using ReLU functions. The computation process involves standard polytope operations, such as intersection and projection, and all of these can be computed by employing sophisticated computational geometry tools, such as MPT3 \cite{MPT3}. The essence of the approach is to be able to obtain an exact output set with respect to the input set.  However, the number of polytopes involved in the computation process increases exponentially with the number of neurons in its worst case performance which makes the method not scalable to neural networks with a large number of neurons.

In practice however, the number of polytopes used in the computation process is usually smaller than worst case operation since empty sets are often produced during the computation procedure. By pre-classifying the active or inactive status of neurons before manipulating the polytopes of each layer and utilizing parallel computing techniques, the computational cost can be reduced to some extent. In the experimental evaluation of these techniques, the authors in \cite{xiang2017reachable_arxiv} were able to decrease the computational time to a third of the original computation time when analyzing  a 7 layer network. Furthermore, the authors were able to utilize this approach for feedback control systems equipped with neural network controllers, where the controlled plant is considered to be a piecewise-linear dynamical system and their efforts are described in \cite{xiang2018reachable_acc}. It should be noted that in the verification of neural network control systems with more general activation functions, it is significantly more difficult. In \cite{scheibler2015towards}, they use bounded model checking (BMC) to create formulas that are solved using the satisfiability modulo theory (SMT)-solver iSAT3, which is able to deal with transcendental functions  such as $\mathtt{exp}$ and $\mathtt{cos}$ that frequently appear in neural network controllers and plants. Although the verification framework is rigorously developed, the verification problem is hardly solved due to the curse of dimensionality and state-space explosion probems.

\begin{figure}
	\begin{center}
	\includegraphics[width=9cm]{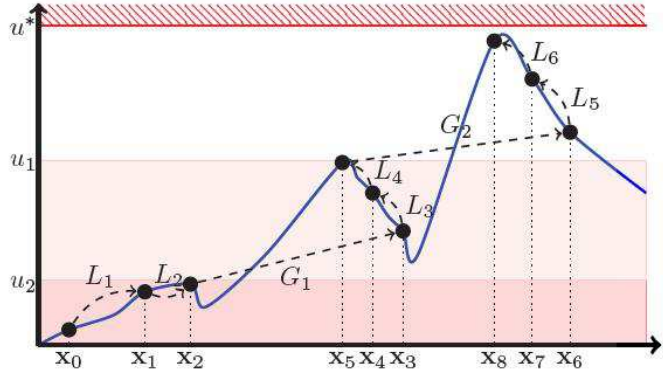}
		\caption{Illustration of a combined local and global search for the MILP in \cite{dutta2017output}. $L_1,\ldots,L_6$ are local searches and $G_1,G_2$ are global searches.}
		\label{fig:MILP_improve}
	\end{center}
\end{figure}

The use of binary variables to encode piecewise linear functions is standard in optimization \cite{Robert2001linear}. In \cite{Lomuscio2017an_arxiv}, the constraints of ReLU functions are encoded as a Mixed-Integer Linear Programming (MILP). Combining output specifications that are expressed in terms of Linear Programming (LP), the verification problem for output set eventually turns to the feasibility problem of MILP. For layer $i$, the MILP encoding is given as
\begin{align}
\mathcal{C}_i = \{&x_{j}^{[i]}\ge W_j^{[i]}x^{[i-1]}+\theta_j^{i}, \nonumber
\\
&x_{j}^{[i]}\le W_j^{[i]}x^{[i-1]}+\theta_j^{i} +M\delta_j^{[i]}, \nonumber
\\
&x_j^{[i]} \ge 0, \nonumber
\\
&x_j^{[i]} \le M(1-\delta_j^{[i]} \mid j =1\ldots\left|L^{[i]}\right|) \}
\end{align}
where $M$ is sufficiently large so that it is larger than the maximum possible output at any node.
Similarly, a MILP problem is formulated in \cite{tjeng2017verifying}, where the authors conduct a robustness analysis and search for adversarial examples in ReLU neural networks. It is well known that MILP is an NP-hard problem and in \cite{dutta2017output,dutta2018output}, the authors elucidate significant efforts  for solving MILP problems efficiently to make the approach scalable. Their methods combine MILP solvers with a local search yielding a more efficient solver for range estimation problems of ReLU neural networks than several other approaches. Basically, a local search is conducted using a gradient search and then a global search is formulated as MILP, as shown in Figure \ref{fig:MILP_improve}. Instead of finding the global optimum directly, it performs the search seeking values greater/smaller than the upper/lower bound obtained in the preceding local search. This is the primary reason for the computational complexity reduction.

\begin{figure}
	\begin{center}
	\includegraphics[width=7cm]{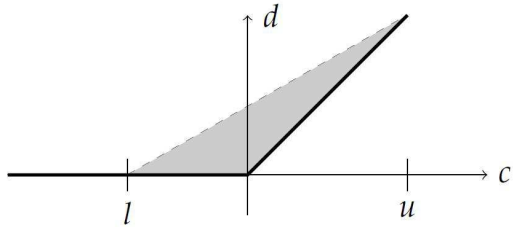}
		\caption{ Figure 1  in \cite{ehlers2017formal}, which illustrates piecewise linear approximation of ReLU functions.}
		\label{fig:PL}
	\end{center}
\end{figure}

Recently, a verification engine for ReLU neural networks called $\mathrm{AI}^2$ was proposed in \cite{GehrAI}. In their approach, the authors abstract perturbed inputs and safety specifications as zonotopes, and reason about their behavior using operations for zonotopes. The framework  $\mathrm{AI}^2$ is capable of handling neural networks of realistic size, and, in particular, their approach has had success dealing with convolutional neural networks. 
In another work, a software tool, called Sherlock, was developed based on the MILP verification approaches \cite{ehlers2017formal}. This LP-based framework combines satisfiability (SAT) solving and linear over-approximation of piecewise linear functions in order to verify ReLU neural networks against convex specifications. An illustration of a piecewise linear approximation for ReLU functions is shown in Figure \ref{fig:PL}. Given the output of a ReLU denoted by $d$ and the input $c \in [l, u]$, the relationship between $c$ and $d$  can be approximated by the linear constraints $d \ge 0$, $d \ge c$, and $d \ge u \frac{c-l}{u-l}$. Based on the LP problem formulation, additional heuristic algorithms were developed to detect infeasibility and imply
phase inference faster.

A tool, named \textit{Planet}, has been developed based on the results in \cite{ehlers2017formal}.  In \cite{katz2017reluplex}, an algorithm, that stems from the Simplex Algorithm for linear functions, for ReLU functions is proposed. Due to the piecewise linear feature of ReLU functions, each node is divided into two nodes. Thus, in their formulation, each node consists of a forward-facing and backward-facing node. If the ReLU semantics are not satisfied, two additional update functions are given to fix the mismatching pairs. Thus, the search process is similar to the Simplex Algorithm that pivots and updates the basic and non-basic variables with the addition of a fixing process for ReLU activation pairs. In their paper, the authors detail an application of their methodology on a deep neural network implementation of a next-generation Airborne Collision Avoidance System for unmanned aircrafts (ACAS-X). A comparison  of the verification approaches mentioned above can be found in \cite{bunel2017piecewise}. Additionally, in the paper, the authors present a novel approach for neural network verification called Branch and Bound Optimization. This approach adds one more layer behind the output layer $cy-b$ to represent the linear property $cy>b$ that we wish to verify. If $cy-b>0$, it means that the property is satisfied, otherwise it is unsatisfiable. Thus, the verification problem is converted into a computation of the minimum or maximum value of the output of the neural network. By treating the neural network as a nonlinear function, model-free optimization methods are utilized to find optimal solution. In order to have a global optimum, the input space is also discretized into sub-regions. This approach is not only applicable to ReLU neural networks, but the model-free method allows the approach to be applied to neural networks with more general activation functions. However, despite its generalization capabilities, in the model-free framework, there is no guarantee that the algorithm will converge to a solution. In \cite{narodytska2017verifying}, a MILP encoding scheme is used for a class of neural networks whose input spaces are encoded as binaries. This MILP encoding  has a similar flavor to the other encodings present in the research literature for non-binarized networks. In their framework, since all the inputs are integer values, the real valued variables can be rounded so that they can be safely removed, resulting in a reformulated integer linear programming (ILP) problem that is smaller in comparison to the original MILP encoding. With the ILP encoding, a SAT solver is utilized in order to reason about the behavior of a mid-size binarized neural network. In another work presented by Osbert Bastani et al. \cite{bastani2016measuring}, the pointwise robustness of a neural network is analyzed using two robustness statistics based on a formulation of robustness as an LP problem. 

Besides the optimization approach proposed in \cite{bunel2017piecewise}, there are few other results for neural networks with general activation functions. In \cite{pulina2010abstraction,pulina2012challenging}, a piecewise-linearization of the nonlinear activation functions is used to reason about their behavior. In this framework, the authors replace the activation functions with piecewise constant approximations and use the bounded model checker hybrid satisfiability (HySAT) \cite{Franzle2007} to analyze various properties. In their paper, the authors highlight the difficulty of scaling this technique and, currently, are only able to tackle small networks with at most 20 hidden nodes. In \cite{xiang2017output}, a simulation-based approach was developed, which used a finite number of simulations/computations to estimate the reachable set of multi-layer neural networks in a general form. Despite this success, the approach lacks the ability to resolve the reachable set computation problem for neural networks that are large-scale, non-convex, and nonlinear. Still, simulation-based approaches, like the one developed in \cite{xiang2017output}, present a plausibly practical and efficient way of reasoning about neural network behavior. The critical step in improving simulation-based approaches is bridging the gap between finitely many simulations and the essentially infinite number of inputs that exist in the continuity set. A  critical concept that is introduced in the work is called maximal sensitivity. Maximal sensitivity, defined formally below, is a measure of the maximal deviation of outputs for a set of inputs suffering disturbances in a bounded cell.
\begin{definition} \label{def}
	Given a neural network $y = f (x)$, an input $x$ and disturbances $\left\|\Delta x \right\|\le\delta$, the maximum sensitivity of the neural network with input error $\delta$ at $x$ is defined by 	
	$
	\epsilon(x,\delta)\triangleq\inf\{\epsilon:~\left\|\Delta y \right\|\le\epsilon, \mathrm{where}~y = f (x)
~\mathrm{and}~\left\|\Delta x \right\|\le\delta\}
$.

\end{definition}

The output set of the neural network can be over-approximated by the union of a finite number of reachtubes computed using a union of individual cells that cover the input set. Thus, verification of a network can be done by checking the existence of intersections of the estimated reachable set and safety regions.  This approach has been extended to allow for the reachable set estimation and verification of Nonlinear Autoregressive-Moving Average (NARMA) models in the form of neural networks \cite{xiang2018reachable}. In particular, it is applicable to a variety of neural networks regardless of the specific form of the activation functions. Given a neural network, there is a trade-off between the precision of the reachable set estimation and the number of simulations used to execute the procedure. In addition, since the approach executes in a layer-by-layer manner, the approximation error will accumulate as the number of layers present in the network increases. In this case, more simulations are required at the expense of increasing the computational cost.  A similar approach for finding adversarial inputs using SMT solvers that relies on a layer-by-layer analysis is presented in \cite{huang2017safety}. The work focuses on the robustness of a neural network where safety is defined in terms of classification invariance within a small neighborhood of one individual input. An exhaustive search of the region is conducted by employing discretization and propagating the analysis layer by layer. In a similar manner, a recent paper, proposed by Wenjie Ruan et al. \cite{ruan2018global}, generalizes the local robustness criterion into a global notion of a set of test examples. In an effort to improve the overall robustness and generalization of neural networks, an empirical study on sensitivity and generalization in neural networks was performed in \cite{novak2018sensitivity}. The experiments, described in this work, survey thousands of models with various fully-connected architectures, optimizers, and other hyper-parameters, as well as four different image classification datasets. The observations contained in this work demonstrate that trained neural networks are more tolerant of input perturbations
in the vicinity of the training data manifold. The factors associated with poor generalization, such as full-batch training or using random labels, correspond to lower robustness, while factors associated with good generalization, such as data
augmentation and ReLU nonlinearities, give rise to more robust functions. 

Another fascinating area in neural network verification is falsification. The main objective of neural network falsification is to find adversarial examples capable of being misclassified that are created by a minimal input perturbation. In this arena, numerous approaches have been proposed with varying levels of efficacy and by conducting these studies the authors seek to improve the overall robustness of trained neural networks. With the above goals in mind, in \cite{huang2017safety}, Xiaowei Huang et al. proposed a framework for verifying the safety of network image classification decisions by searching for adversarial examples within a specified region. Their methods are implemented in a software tool called DLV (Deep Learning Verification). The authors cite that if a misclassification exists within a region, it will be found. The essence of their approach stems from the idea of point-wise robustness and, in order for a network to be considered safe, it should be stable in the presence of slight input perturbations. Thus, their method relies on propagating search regions and manipulations throughout the network, layer by layer. In a similar work by Chih-Hong Chen et al. \cite{cheng2017maximum}, the authors study the problem of ensuring that network based classifiers are resilient towards noisy or maliciously manipulated sensory input. Thus, they define resilience as the maximum amount of input perturbation that can be tolerated by the network and encode their problem using MILP. Their work is distinguishable since their methods provide measures that are a property of the network rather than related to a single input image. Thus, they allow others to quantitatively and formally gauge the resilience of a network.
\par Motivated by the wealth of adversarial example generation approaches for neural network image classifiers, such as in \cite{novak2018sensitivity,weng2018towards,2018arXiv180406760E, DongDiscovering2017, hein2017formal}, Tommasso Derossi argued that majority of current techniques neglect the semantics and context of the overall systems in which the neural network is deployed. Moreover, the author argues that neglecting these aspects is ill-advised since context plays a crucial role in safety assurance and verification. Thus, in their paper \cite{dreossi2018semantic}, the authors present several ideas for integrating semantics into adversarial learning, including a semantic modification space and the use of more detailed information about the outputs produced by machine learning models. In another interesting work \cite{wicker2018feature}, Matthew Wicker et al. propose a feature-guided approach for testing image classifiers that does not require architectural knowledge of the network at hand. Their scheme is based on object detection techniques and they formulate the creation of adversarial examples as a two player stochastic game using Monte Carlo tree searching to explore the game state space in order to search for adversarial examples. The evaluation of their  techniques is promising and the authors demonstrate that they can provide conditions in which no adversarial example exists. In a similar work by Tsui Weng et al. \cite{2018arXiv180110578W}, they present an attack independent robustness metric against adversarial examples for neural networks. Their approach converts the robustness analysis into a local Lipschitz constant estimation problem and uses Extreme Value Theory for efficient solving.

\subsection{Testing for Adversarial Machine Learning Examples}
Equally important are the numerous proposed defenses against adversarial manipulation and testing suites for neural network based classifiers. In \cite{kolter2017provable}, the authors present a method for training deep ReLU based classifiers that are provably robust against norm-bounded adversarial examples. The authors cite that their methods are able to detect all adversarial examples from a specified input set. However, their methods are overly conservative and may mistakenly flag inputs that are not adversarial. In another work, Aditi Raghunathan et al. \cite{raghunathan2018certified} propose defenses based on regularization for networks with a single hidden layer. Their approach is based on a semidefinite relaxation that outputs a differentiable certificate that no adversarial examples exist. The certificate is defined by computing the worst-case loss for a one layer hidden network. In \cite{SinhaCertifying2017}, the authors present a framework for defending against adversarial examples by using a distributionally robust optimization that guarantees performance through the use of a Lagrangian penalty formulation of the underlying training data distribution. They also provide a training method that augments parameters with worst-case distortions of the training data. The authors argue that their methods achieve satisfactory levels of robustness with little associated computational and statistical cost. Complementary to these techniques are the testing suites proposed to assess neural network safety. In \cite{sun2018testing}, the authors propose a set of four test coverage criteria for deep learning systems inspired by the modified condition/decision coverage criterion developed by NASA for avionics software. Their methods are implemented in a software tool  called DeepCover.While their methods do not formally guarantee correct functionality, they give users confidence that the networks are reasonably free from defect. In a similar work by Lei Ma et al. \cite{LeiMaDeepGauge2018}, the authors present DeepGuage, a tool implementing multi-granularity testing criteria for deep learning systems. The aim of their tool is the construction of high quality test data that will provide guarantees about the performance, robustness, and generality of deep learning systems. Similarly, in \cite{tian2017deeptest}, an automatic test case generator is presented that leverages real-world changes in driving conditions like rain, fog, lighting conditions, etc. The tool, called \textit{DeepTest}\endnote{https://deeplearningtest.github.io/deepTest/}, systematically explores different parts of the deep neural network logic by generating test inputs that maximize the number of activated neurons. An improved version of the tool, called DeepXplore, is proposed in \cite{pei2017deepxplore}, which is the first efficient whitebox testing framework for large-scale deep learning systems. 

\subsection{Other Approaches}

In \cite{dreossi2017compositional}, a compositional falsification framework where a temporal logic falsifier and a machine learning analyzer cooperate with the aim of finding falsifying  executions of CPS models that involve machine learning components is presented.

In \cite{Lomuscio2017an_arxiv}, Alessio Lomuscio and Lalit Maganti present a linear programming based method for reachability analysis in feed-forward neural networks with ReLU activation functions. Specifically, they address whether an arbitrary output can be gained for a fixed set of inputs by encoding the neural network into a linear constraint system. The authors then tailor the objective function to ensure safe tolerances in the floating point arithmetic associated with linear programming. Once the encoding of the network has been completed, the authors combine a set of specifications with the linear system and use Gurobi to achieve a solution. With the solutions returned by Gurobi, users can gauge the correctness of the network analyzed. The authors include an experimental analysis of several networks for various problems in control theory and demonstrate that their method solves reachability problems for networks with sizes on the order of 5000 neurons. 

In this paper \cite{scheibler2015towards}, the authors study the safety verification of a typical control problem, called the Cart Pole System, using bounded model checking. The authors use to the SMT solver, iSAT3, to check satisfiability of the created formulas. Their experimental results demonstrate that even the smallest bounded model checking instances could hardly be solved due to the non-linearity and non-invertibility of neural networks. However, in this work, the authors do not solve the safety verification problem.

In this paper \cite{10.1007/978-3-540-30138-7_3}, Zeshan Kurd et al. define a constrained type of artificial neural network (ANN) that enables analytical certification arguments while still possessing profitable performance attributes. The proposed model is based upon a fuzzy system called the Fuzzy Self-Organising Map (FSOM). This model is a type of hybrid ANN with behavior that can be qualitatively and quantitatively described. Thus, one can reason about the safety of such a network by analyzing its adherence to safety requirements expressed as safety constraints. In their paper, the authors argue that this framework allows for the construction of arguments that highlight potential failure modes in the FSOM and, therefore, it can be considered a Safety Critical Artificial Neural network. Additionally, the authors demonstrate that the FSOM can be used for nonlinear function approximation with enhanced generalization. While most methods related to ANN verification interpret ANN as black boxes, this framework allows for white box analysis.

In their paper \cite{PulinaNever}, Luca Pulina and Armando Tacchella present an abstraction-refinement and satisfiability modulo theory based tool for verifying feed-forward neural networks. Their scheme is based on encoding the  network into a boolean satisfaction problem over linear arithmetic constraints. The inputs to the tool are a training set, several parameters concerning the structure of the network, a set of safety thresholds, an abstraction parameter, and a refinement rate. Using these inputs, the tool seeks to verify the safety thresholds. If it generates a counterexample, it passes the example to the original network in order to identify spurious counterexamples. Otherwise, the tool either returns that the network is safe or a counterexample and seeks to improve the network in an automated fashion. The authors evaluated their tool on various case studies in order to demonstrate the success and scalability of the approach.

The paper, \cite{dvijotham2018dual}, presents a novel approach for neural network verification based on optimization and duality. The verification problem is posed as an optimization problem that tries to find the largest violation of a property related to the output  of the network. If the largest violation is smaller than zero, then the property is verified. To do this, the authors use ideas from duality in optimization to obtain bounds on the optimal value of the verification problem. The authors cite that their method is sound but unfortunately in some cases cannot prove that a property is true. Their approach works on feedforward neural networks of any architecture and with arbitrary activation functions. Furthermore, it has been demonstrated that it performs just as well as the other verification methods that are currently available for networks with piecewise linear activation functions. Their approach is anytime in that it can be stopped at any point with a valid bound on the verification objective.

The following paper, \cite{SunConcolic2018}, explores concolic testing for deep neural networks. Concolic Testing explores the execution paths of a software program to increase code coverage by alternating between concrete program execution and symbolic analysis. In this paper, the authors use quantified linear arithmetic over rationals (QLAR) to express test requirements aimed at finding adversarial examples. QLAR expresses a set of safety-related properties including Lipchitz continuity and several other coverage properties. The method proposed by the authors alternates between evaluating a DNN's activation patterns and symbolically generating new inputs. Their experimental results demonstrate that their methodology is superior to DeepXplore. Their method is implemented in a software tool.

The following paper \cite{XieFuzzing2018} presents an automated testing framework called DeepHunter for identifying potential defects in deep neural networks using coverage-guided fuzz (CGF) testing. In this framework, the authors systematically create mutations of inputs in order to generate a set of tests that will test a neural network's behavior in corner case scenarios. The creation of input mutations is guided by a fuzzing strategy and one of the key components of CGF is feedback guidance. Thus, CGF is a feedback-guided framework that consistently updates the fuzzing strategy in order to promote efficiency and maximize test coverage. Additionally, the authors cite that their approach is scalable to real world neural networks since their approach organizes tests in batches and prioritizes tests that are given by the feedback signal. The authors evaluate their approach on seven neural networks with varying complexities and demonstrate that their approach can successfully detect erroneous behavior and potential defects in deep neural networks.

Recently, numerous frameworks for deep neural network verification have made use of SMT solvers. However, the majority of these approaches are limited by the high computational cost caused by using these solvers. In this paper \cite{WangFormal2018}, the authors present  a method for formally checking security properties of neural networks by leveraging interval arithmetic to compute rigorous bounds on the DNN's outputs. Their approach is easily parallelizable and makes use of symbolic interval analysis in order to minimize overestimations. The authors implement their approach as part of ReluVal, a system for checking the security properties of Relu-based DNNs. Their approach has been demonstrated to be able to outperform Reluplex by 200 times. Given an input range X and security property P, ReluVal propagates it layer by layer and applies a variety of optimizations to calculate the output range.

In this paper \cite{DBLP:journals/corr/abs-1710-03107}, Chih-Hong Cheng et al. present a study of the verification of Binarizied neural networks (BNNs). This class of network has been demonstrated to be more power efficient in embedded system devices due to the fact that their weights, inputs, intermediate signals, outputs, and activation constraints are binary valued. Thus, the forward propagation of input signals is reduced to bit arithmetic. The authors argue that the verification of BNNs can be reduced to hardware verification and  represents a more scalable problem than traditional neural network verification. The authors build on well known methods from hardware verification and transform the BNN and its specification into a combinational miter, which is an encoded equivalence check of two boolean circuits. This is then transformed into a SAT problem which can be solved using a variety of available software tools. The authors were not able to verify a network with a thousand nodes and cite that their future work will focus on verifying networks currently present in embedded devices.

The following paper, \cite{hull2002verification}, presents a method for obtaining formal guarantees on the robustness of a classifier by giving instance-specific lower bounds on the norm of the input manipulation required to change the classifier decision. Their techniques can be applied to two widely used families of classifiers: kernel methods and neural networks. Furthermore, they propose the Cross-Lipschitz regularization functional that improves upon both the formal guarantees of the resulting classifier while achieving similar prediction performances as other forms of regularization. The authors also provide algorithms based on the first order approximation of the classifier with generated adversarial examples satisfying box constraints.

The following paper, \cite{JI20177158}, presents the problem of estimation of reachable sets for Markov jump bidirectional associative memory (BAM) neural networks. The neural networks have inputs that are either unit-energy or unit-peak bounded disturbances and have time-varying delay and transition rates. The technique of partitioning the delays is used in response to the significant conservatism of the solution. They also focus on Markov jump BAM neural networks that are uncertain and have transition probabilities that are unknown. Further work could include estimation of the reachable set for Markov jump systems that are uncertain.

The following paper, \cite{ZakrzewskiRandomized2004}, presents a randomized approach for rigorously verifying neural networks in safety critical applications. In an effort to mitigate challenges related to the curse of dimensionality, the authors make use of Monte Carlo methods to estimate the probability of neural network failure. However, although Monte Carlo methods are more efficient than methods that deterministically search through hyper-rectangular input spaces, they are probabilistic in nature. The objective of their approach is to verify that the input and output error of a network is small and remains within a specified bound. By drawing from sufficiently many points from an n-dimensional rectangle, the probability of the error becoming large may be made arbitrarily small. The authors further demonstrate that although the number of samples needed to guarantee this may be large, it is not as prohibitive as other methods.

\def\networkHiddenNeuronTotal{N}
\def\networkLayers{L}
\def\networkInputs{$I_N$}
\def\networkOutputs{$O_N$}
	
\noindent Breakdown of the neural network verification software tools currently available in the research literature. 

Table \ref{NNVerificationTable}, Table \ref{NNAdversarial}, and Table \ref{NNCoverTable} summarize the available tools for neural network verification, adversarial input robustness verification, and neural network testing respectively. They list key characteristics about each tool, reference the papers in which the methods can be found, and also provide the URL where each tool can be found. In each table, the letters \textit{I,O,L} and \textit{N} denote the number of inputs, outputs, layers, and neurons respectively. The analysis time listed is the worst case execution time of the verified examples in each paper.
\begin{center}
	\begin{longtable}{  p{0.20\textwidth} | p{0.10\textwidth} | p{0.15\textwidth}  | p{0.17\textwidth} | p{0.20\textwidth} | p{0.10\textwidth} } 
	\caption{Neural Network Verification Tools \label{NNVerificationTable}}\\
			Tool Name & Network Type & Verification Approach & Network Size (I,O,L,N) & Experimental Setup & Analysis Time (s)\\[0.5cm]
			\hline
			Reluplex\endnote{https://github.com/guykatzz/ReluplexCav2017} \cite{katz2017reluplex} & FNN & SMT & (5,5,8,2400) & \scriptsize{Ubuntu 16.04} & 394517 \\
            Sherlock\endnote{https://github.com/souradeep-111/sherlock} \cite{dutta2017output} & FNN & MIP & (4,3,24,3822) & \scriptsize{Linux Server, Ubuntu 17.04,24 cores,64GB RAM} & 261540 \\
            AnalyzeNN\endnote{https://github.com/dreossi/analyzeNN} \cite{dreossi2017compositional} &  FNN, CNN & Signal Temporal Logic Falsification   & (1,1,8,261)& \scriptsize{Dell XPS 8900 Intel(R) Core(TM) i7-6700 CPU 3.40 GHz 16 GB RAM, GPU NVIDIA GeForce GTX Titan X} & N/A \\
            $\textrm{AI}^2$ \endnote{http://ai2.ethz.ch/} \cite{GehrAI} & FNN, CNN & Abstract Interpretation & (1,1,9,53000) & \scriptsize{Ubuntu 16.04.3 LTS, 2 Intel Xeon E5-2690 Processors, 512 GB RAM} & 75-3600 \\
            PLNN\endnote{https://github.com/oval-group/PLNN-verification} \cite{bunel2017piecewise} & FNN, CNN & Branch and Bound & (5,5,6,300) & \scriptsize{single core,  i7-5930K CPU, 32GB RAM, Timeout of 2 hrs, maximum memory 20 GB} & 5248.1 \\
            Planet\endnote{https://github.com/progirep/planet}  \cite{ehlers2017formal} & FNN, CNN & SAT,LP & (1,1,3,1341) & \scriptsize{Linux Intel Core i5-4200U 1.60 GHz CPU, 8 GB RAM, 1 hour timeout} &1955 \\
            NeVer\endnote{http://www.mind-lab.it/never} \cite{PulinaNever} & FNN & SMT & (1,1,1,32) & \scriptsize{10 Linux workstations,Intel Core 2 Duo 2.13 GHz PCs with 4GB of RAM, Linux Debian 2.6.18.5} & 344.59 \\
            \hline
    \end{longtable}
\end{center}

\begin{center}
	\begin{longtable}{  p{0.20\textwidth} | p{0.10\textwidth} | p{0.15\textwidth}  | p{0.17\textwidth} | p{0.20\textwidth} | p{0.10\textwidth} } 
	\caption{Verification Tools Concerned with Adversarial Input Robustness \label{NNAdversarial}}\\
			Tool Name & Network Type & Verification Approach & Network Size (I,O,L,N) & Experimental Setup & Analysis Time (s)\\[0.1cm]
			\hline
            VeriDeep/DLV\endnote{https://github.com/VeriDeep/DLV}\cite{huang2017safety}  & FNN, CNN & SMT & \small{(1,1,16, 138,357,544)} parameters &\scriptsize{MacBook Pro laptop, 2.7 GHz Intel Core
            i5 CPU, 8 GB RAM} & 60-120 per input \\
            DeepGo\endnote{https://github.com/TrustAI/DeepGO} \cite{RuanDeepGO2018} & FNN, CNN & \small{Global Nested Optimization} & (1,1,7,412) & \scriptsize{Notebook computer Matlab 2018a i7-7700HQ CPU and 16GB RAM } & 5 \\
            $L_0$-TRE\endnote{https://github.com/L0-TRE/L0-TRE} \cite{ruan2018global} & FNN, CNN & $L_0$-Norm robustness evaluation & \small{(1,1,101, 44,654,504 parameters)} & \scriptsize
            {Ubuntu 14.04.3 LTS NVIDIA GeForce GTX TITAN Black Intel(R) Core(TM) i5-4690s 3.20 GHz x 4} & \footnotesize{anytime algorithm} \\
            SafeCV\endnote{https://github.com/matthewwicker/SafeCV} \cite{wicker2018feature} &  CNN & \small{Adversarial Black-Box Falsification} & \small{(1,1,12,250,858 parameters)} & \small{N/A} & 20 \\
            Certified ReLU Robustness\endnote{https://github.com/huanzhang12/CertifiedReLURobustness} \cite{weng2018towards} & FNN, CNN & MIP & \small{(1,1,7, 10,000)} & \scriptsize{a Intel Xeon E5-2683v3 (2.0 GHz) CPU} & 858 \\
            NNAF\endnote{https://github.com/Microsoft/NeuralNetworkAnalysis} \cite{bastani2016measuring} & FNN, CNN & LP & \small{(1,1,7,60000 parameters)} & \small{8 core CPU} & $1.5$ per input \\
            Convex Adversarial\endnote{https://github.com/locuslab/convex\_adversarial}\cite{kolter2017provable} & FNN, CNN & \small{Convex Outer Approximation} & (2,2,6,404) & \scriptsize{NVIDIA Titan X GPU} & 18000 \\
            \hline
    \end{longtable}	
\end{center}

\begin{center}
	\begin{longtable}{  p{0.20\textwidth} | p{0.10\textwidth} | p{0.15\textwidth}  | p{0.17\textwidth} | p{0.18\textwidth} | p{0.10\textwidth} } 
	\caption{Neural Network Testing Tools \label{NNCoverTable}}\\
			Tool Name & Network Type & Verification Approach & Network Size (I,O,L,N) & Experimental Setup & Analysis Time (s)\\
			\hline
			DeepCover\endnote{https://github.com/deep-cover/deep-cover} \cite{sun2018testing} & FNN, CNN & White-box LP inspired test criteria & (1,1,4,14208) & \scriptsize{Macbook 2.5 GHz Intel Core i5 8 GB RAM} & \footnotesize{0.94/LP call}\\[0.8 cm]
			DeepXplore\endnote{https://github.com/peikexin9/deepxplore} \cite{pei2017deepxplore} & FNN, CNN & Test input generation & (1,1,50,94059) & \scriptsize{Linux Ubuntu 16.04, Intel i7-6700HQ 2.60GHz, 4 cores, 16GB RAM NVIDIA GTX 1070 GPU} &\small{196.4 for 100\% coverage} \\[0.5cm]
			 DeepConcolic\endnote{https://github.com/TrustAI/DeepConcolic} \cite{SunConcolic2018} & FNN, CNN & \small{Quantified Linear Arithmetic concolic testing} & (1,1,10,538) & \scriptsize{Intel(R) Core (TM) i5-4690S CPU @ 3.20 GHz x 4} & N/A \\
			 \hline
			
    \end{longtable}
\end{center}

%% file: conclusion.tex
\section{Conclusion}
This paper surveys formal methods, verification and validation (V\&V), and architectural approaches to analyze and ideally assure safety of cyber-physical systems (CPS) that incorporate learning enabled components (LECs), such as neural networks.

%

%% file: main.bbl
\begin{thebibliography}{100}

\bibitem{AartsLearning2012}
Fides Aarts, Harco Kuppens, Jan Tretmans, Frits Vaandrager, and Sicco Verwer.
\newblock {Learning and Testing the Bounded Retransmission Protocol}.
\newblock In Jeffrey Heinz, Colin Higuera, and Tim Oates, editors, {\em
  Proceedings of the Eleventh International Conference on Grammatical
  Inference}, volume~21 of {\em Proceedings of Machine Learning Research},
  pages 4--18, University of Maryland, College Park, MD, USA, 05--08 Sep 2012.
  PMLR.

\bibitem{AkametaluReachability2014}
A.~K. Akametalu, J.~F. Fisac, J.~H. Gillula, S.~Kaynama, M.~N. Zeilinger, and
  C.~J. Tomlin.
\newblock {Reachability-Based Safe Learning with Gaussian Processes}.
\newblock In {\em 53rd IEEE Conference on Decision and Control}, pages
  1424--1431, Dec 2014.

\bibitem{wrro127573}
Robert~David Alexander, Rob Ashmore, and Andrew Banks.
\newblock {The State of Solutions for Autonomous Systems Safety}.
\newblock February 2018.

\bibitem{AlshiekhSafeRL2017}
Mohammed Alshiekh, Roderick Bloem, R{\"{u}}diger Ehlers, Bettina
  K{\"{o}}nighofer, Scott Niekum, and Ufuk Topcu.
\newblock {Safe Reinforcement Learning via Shielding}.
\newblock {\em CoRR}, abs/1708.08611, 2017.

\bibitem{AngluinQuery1988}
D.~Angluin.
\newblock {Queries and Concept Learning}.
\newblock In {\em Machine Learning}, pages 319--342. Springer, 1988.

\bibitem{AnsinAutomated2013}
Rasmus Ansin and Didrik Lundberg.
\newblock {\em {Automated Inference of Excitable Cell Models as Hybrid
  Automata}}.
\newblock PhD thesis, 2013.

\bibitem{Arulkumaran2017}
K.~{Arulkumaran}, M.~P. {Deisenroth}, M.~{Brundage}, and A.~A. {Bharath}.
\newblock {A Brief Survey of Deep Reinforcement Learning}.
\newblock {\em ArXiv e-prints}, August 2017.

\bibitem{asarin2011parametric}
Eugene Asarin, Alexandre Donz{\'e}, Oded Maler, and Dejan Nickovic.
\newblock Parametric identification of temporal properties.
\newblock In {\em International Conference on Runtime Verification}, pages
  147--160. Springer, 2011.

\bibitem{AslundVirtuouslySafeRL2018}
H.~{Aslund}, E.~{Mahdi El Mhamdi}, R.~{Guerraoui}, and A.~{Maurer}.
\newblock {Virtuously Safe Reinforcement Learning}.
\newblock {\em ArXiv e-prints}, May 2018.

\bibitem{bak2015rtss}
Stanley Bak and Taylor~T. Johnson.
\newblock {Periodically-Scheduled Controller Analysis using Hybrid Systems
  Reachability and Continuization}.
\newblock In {\em 36th IEEE Real-Time Systems Symposium (RTSS 2015)}, San
  Antonio, Texas, December 2015. IEEE Computer Society.

\bibitem{johnson2016tecs}
Stanley Bak, Taylor~T Johnson, Marco Caccamo, and Lui Sha.
\newblock {Real-Time Reachability for Verified Simplex Design}.
\newblock In {\em 2014 IEEE Real-Time Systems Symposium}, pages 138--148, Dec
  2014.

\bibitem{BalkanUnderminer2017}
Ayca Balkan, Paulo Tabuada, Jyotirmoy~V. Deshmukh, Xiaoqing Jin, and James
  Kapinski.
\newblock {Underminer: A Framework for Automatically Identifying Nonconverging
  Behaviors in Black-Box System Models}.
\newblock {\em ACM Trans. Embed. Comput. Syst.}, 17(1):20:1--20:28, December
  2017.

\bibitem{bastani2016measuring}
Osbert Bastani, Yani Ioannou, Leonidas Lampropoulos, Dimitrios Vytiniotis,
  Aditya~V. Nori, and Antonio Criminisi.
\newblock {Measuring Neural Net Robustness with Constraints}.
\newblock {\em CoRR}, abs/1605.07262, 2016.

\bibitem{1866}
B.~Bavarian.
\newblock {Introduction to Neural Networks for Intelligent Control}.
\newblock {\em IEEE Control Systems Magazine}, 8(2):3--7, April 1988.

\bibitem{BerkenkampSafeModelRL2017}
Felix Berkenkamp, Matteo Turchetta, Angela Schoellig, and Andreas Krause.
\newblock {Safe Model-based Reinforcement Learning with Stability Guarantees}.
\newblock In I.~Guyon, U.~V. Luxburg, S.~Bengio, H.~Wallach, R.~Fergus,
  S.~Vishwanathan, and R.~Garnett, editors, {\em Advances in Neural Information
  Processing Systems 30}, pages 908--918. Curran Associates, Inc., 2017.

\bibitem{bertsekas2005dynamic}
Dimitri~P Bertsekas, Dimitri~P Bertsekas, Dimitri~P Bertsekas, and Dimitri~P
  Bertsekas.
\newblock {\em Dynamic programming and optimal control}, volume~1.
\newblock Athena scientific Belmont, MA, 2005.

\bibitem{BogomolovPDDL+2015}
Sergiy Bogomolov, Daniele Magazzeni, Stefano Minopoli, and Martin Wehrle.
\newblock {PDDL+ Planning with Hybrid Automata: Foundations of Translating Must
  Behavior}.
\newblock 2015.

\bibitem{bojarski2016end}
M.~Bojarski, D.~Del~Testa, et~al.
\newblock {End to End Learning for Self-Driving Cars}.
\newblock {\em arXiv preprint arXiv:1604.07316}, 2016.

\bibitem{bunel2017piecewise}
Rudy Bunel, Ilker Turkaslan, Philip~HS Torr, Pushmeet Kohli, and M~Pawan Kumar.
\newblock {Piecewise Linear Neural Network verification: A comparative study}.
\newblock {\em arXiv preprint arXiv:1711.00455}, 2017.

\bibitem{7067402}
B.~Chen, H.~Zhang, and C.~Lin.
\newblock {Observer-Based Adaptive Neural Network Control for Nonlinear Systems
  in Nonstrict-Feedback Form}.
\newblock {\em IEEE Transactions on Neural Networks and Learning Systems},
  27(1):89--98, Jan 2016.

\bibitem{6627983}
C.~L.~P. Chen, Y.~J. Liu, and G.~X. Wen.
\newblock {Fuzzy Neural Network-Based Adaptive Control for a Class of Uncertain
  Nonlinear Stochastic Systems}.
\newblock {\em IEEE Transactions on Cybernetics}, 44(5):583--593, May 2014.

\bibitem{ChenDeepDriving}
Chenyi Chen, Ari Seff, Alain Kornhauser, and Jianxiong Xiao.
\newblock Deepdriving: Learning affordance for direct perception in autonomous
  driving.
\newblock In {\em Proceedings of the 2015 IEEE International Conference on
  Computer Vision (ICCV)}, ICCV '15, pages 2722--2730, Washington, DC, USA,
  2015. IEEE Computer Society.

\bibitem{5430947}
M.~Chen, S.~S. Ge, and B.~V.~E. How.
\newblock {Robust Adaptive Neural Network Control for a Class of Uncertain MIMO
  Nonlinear Systems With Input Nonlinearities}.
\newblock {\em IEEE Transactions on Neural Networks}, 21(5):796--812, May 2010.

\bibitem{ChenUAVSafety2017}
Mo~Chen, Qie Hu, Jaime~F. Fisac, Kene Akametalu, Casey Mackin, and Claire~J.
  Tomlin.
\newblock {Reachability-Based Safety and Goal Satisfaction of Unmanned Aerial
  Platoons on Air Highways}.
\newblock {\em CoRR}, abs/1602.08150, 2016.

\bibitem{ChenSafePlatooning2017}
Mo~Chen, Qie Hu, Casey Mackin, Jaime~F. Fisac, and Claire~J. Tomlin.
\newblock {Safe Platooning of Unmanned Aerial Vehicles via Reachability}.
\newblock {\em CoRR}, abs/1503.07253, 2015.

\bibitem{cheng2017maximum}
Chih-Hong Cheng, Georg N{\"u}hrenberg, and Harald Ruess.
\newblock {Maximum Resilience of Artificial Neural Networks}.
\newblock In {\em International Symposium on Automated Technology for
  Verification and Analysis}, pages 251--268. Springer, 2017.

\bibitem{DBLP:journals/corr/abs-1710-03107}
Chih{-}Hong Cheng, Georg N{\"{u}}hrenberg, and Harald Ruess.
\newblock {Verification of Binarized Neural Networks}.
\newblock {\em CoRR}, abs/1710.03107, 2017.

\bibitem{7451284}
Z.~Chu, D.~Zhu, and S.~X. Yang.
\newblock {Observer-Based Adaptive Neural Network Trajectory Tracking Control
  for Remotely Operated Vehicle}.
\newblock {\em IEEE Transactions on Neural Networks and Learning Systems},
  28(7):1633--1645, July 2017.

\bibitem{4812095}
S.~Cong and Y.~Liang.
\newblock {PID-Like Neural Network Nonlinear Adaptive Control for Uncertain
  Multivariable Motion Control Systems}.
\newblock {\em IEEE Transactions on Industrial Electronics}, 56(10):3872--3879,
  Oct 2009.

\bibitem{6514578}
S.~L. Dai, C.~Wang, and M.~Wang.
\newblock {Dynamic Learning From Adaptive Neural Network Control of a Class of
  Nonaffine Nonlinear Systems}.
\newblock {\em IEEE Transactions on Neural Networks and Learning Systems},
  25(1):111--123, Jan 2014.

\bibitem{DongDiscovering2017}
Yinpeng Dong, Fangzhou Liao, Tianyu Pang, Xiaolin Hu, and Jun Zhu.
\newblock {Discovering Adversarial Examples with Momentum}.
\newblock {\em CoRR}, abs/1710.06081, 2017.

\bibitem{donze2010breach}
Alexandre Donz{\'e}.
\newblock Breach, a toolbox for verification and parameter synthesis of hybrid
  systems.
\newblock In {\em International Conference on Computer Aided Verification},
  pages 167--170. Springer, 2010.

\bibitem{dreossi2018semantic}
T.~{Dreossi}, S.~{Jha}, and S.~A. {Seshia}.
\newblock {Semantic Adversarial Deep Learning}.
\newblock {\em ArXiv e-prints}, April 2018.

\bibitem{dreossi2017compositional}
Tommaso Dreossi, Alexandre Donz{\'e}, and Sanjit~A Seshia.
\newblock {Compositional Falsification of Cyber-Physical Systems with Machine
  Learning Components}.
\newblock In {\em NASA Formal Methods Symposium}, pages 357--372. Springer,
  2017.

\bibitem{dutta2017output}
Souradeep Dutta, Susmit Jha, Sriram Sanakaranarayanan, and Ashish Tiwari.
\newblock {Output Range Analysis for Deep Neural Networks}.
\newblock {\em arXiv preprint arXiv:1709.09130}, 2017.

\bibitem{dutta2018output}
Souradeep Dutta, Susmit Jha, Sriram Sankaranarayanan, and Ashish Tiwari.
\newblock {Output Range Analysis for Deep Feedforward Neural Networks}.
\newblock In Aaron Dutle, C{\'e}sar Mu{\~{n}}oz, and Anthony Narkawicz,
  editors, {\em NASA Formal Methods}, pages 121--138, Cham, 2018. Springer
  International Publishing.

\bibitem{dvijotham2018dual}
Krishnamurthy Dvijotham, Robert Stanforth, Sven Gowal, Timothy Mann, and
  Pushmeet Kohli.
\newblock {A Dual Approach to Scalable Verification of Deep Networks}.
\newblock {\em arXiv preprint arXiv:1803.06567}, 2018.

\bibitem{ehlers2017formal}
R{\"{u}}diger Ehlers.
\newblock {Formal Verification of Piece-Wise Linear Feed-Forward Neural
  Networks}.
\newblock {\em CoRR}, abs/1705.01320, 2017.

\bibitem{FisacSafety2017}
Jaime~F. Fisac, Anayo~K. Akametalu, Melanie~Nicole Zeilinger, Shahab Kaynama,
  Jeremy~H. Gillula, and Claire~J. Tomlin.
\newblock {A General Safety Framework for Learning-Based Control in Uncertain
  Robotic Systems}.
\newblock {\em CoRR}, abs/1705.01292, 2017.

\bibitem{FisacPragmatic2017}
Jaime~F. Fisac, Monica~A. Gates, Jessica~B. Hamrick, Chang Liu, Dylan
  Hadfield{-}Menell, Malayandi Palaniappan, Dhruv Malik, S.~Shankar Sastry,
  Thomas~L. Griffiths, and Anca~D. Dragan.
\newblock {Pragmatic-Pedagogic Value Alignment}.
\newblock {\em CoRR}, abs/1707.06354, 2017.

\bibitem{FiterauCombining2016}
Paul Fiter{\u{a}}u-Bro{\c{s}}tean, Ramon Janssen, and Frits Vaandrager.
\newblock {Combining Model Learning and Model Checking to Analyze TCP
  Implementations}.
\newblock In Swarat Chaudhuri and Azadeh Farzan, editors, {\em Computer Aided
  Verification}, pages 454--471, Cham, 2016. Springer International Publishing.

\bibitem{Franzle2007}
Martin Fr{\"a}nzle and Christian Herde.
\newblock {HySAT: An efficient proof engine for bounded model checking of
  hybrid systems}.
\newblock {\em Formal Methods in System Design}, 30(3):179--198, Jun 2007.

\bibitem{170966}
T.~Fukuda and T.~Shibata.
\newblock {Theory and Applications of Neural Networks for Industrial Control
  Systems}.
\newblock {\em IEEE Transactions on Industrial Electronics}, 39(6):472--489,
  Dec 1992.

\bibitem{FultonSafe18}
Nathan Fulton and Andr{\'e} Platzer.
\newblock {Safe Reinforcement Learning via Formal Methods: Toward Safe Control
  Through Proof and Learning}.
\newblock In Sheila McIlraith and Kilian Weinberger, editors, {\em Proceedings
  of the Thirty-Second {AAAI} Conference on Artificial Intelligence, February
  2-7, 2018, New Orleans, Louisiana, {USA.}} {AAAI} Press, 2018.

\bibitem{GarciaSurvey}
Javier Garc{{\'i}}a and Fernando Fern{{\'a}}ndez.
\newblock {A Comprehensive Survey on Safe Reinforcement Learning}.
\newblock {\em Journal of Machine Learning Research}, 16:1437--1480, 2015.

\bibitem{1215406}
S.~S. Ge and Jin Zhang.
\newblock {Neural-Network Control of Nonaffine Nonlinear System With Zero
  Dynamics by State and Output Feedback}.
\newblock {\em IEEE Transactions on Neural Networks}, 14(4):900--918, July
  2003.

\bibitem{GehrAI}
Timon Gehr, Matthew Mirman, Dana~Draschler Cohen, Petar Tsankov, Swarat
  Chaudhuri, and Martin Vechev.
\newblock {$AI^2$: Safety and Robustness Certification of Neural Networks with
  Abstract Interpretation}.
\newblock {\em IEEE Symposium on Security and Privacy}, 39, May 2018.

\bibitem{GiantamidisLearning2016}
Georgios Giantamidis and Stavros Tripakis.
\newblock {Learning Moore Machines from Input-Output Traces}.
\newblock In John Fitzgerald, Constance Heitmeyer, Stefania Gnesi, and Anna
  Philippou, editors, {\em FM 2016: Formal Methods}, pages 291--309, Cham,
  2016. Springer International Publishing.

\bibitem{GillulaGuaranteedSafeRL2012}
J.~H. Gillula and C.~J. Tomlin.
\newblock {Guaranteed Safe Online Learning via Reachability: Tracking a Ground
  Target Using a Quadrotor}.
\newblock In {\em 2012 IEEE International Conference on Robotics and
  Automation}, pages 2723--2730, May 2012.

\bibitem{grosu2007learning}
Radu Grosu, Sayan Mitra, Pei Ye, Emilia Entcheva, IV~Ramakrishnan, and Scott~A
  Smolka.
\newblock Learning cycle-linear hybrid automata for excitable cells.
\newblock In {\em International Workshop on Hybrid Systems: Computation and
  Control}, pages 245--258. Springer, 2007.

\bibitem{1281757}
P.~Gupta and J.~Schumann.
\newblock {A Tool for Verification and Validation of Neural Network Based
  Adaptive Controllers for High Assurance Systems}.
\newblock In {\em Eighth IEEE International Symposium on High Assurance Systems
  Engineering, 2004. Proceedings.}, pages 277--278, March 2004.

\bibitem{7113913}
W.~He, Y.~Dong, and C.~Sun.
\newblock {Adaptive Neural Impedance Control of a Robotic Manipulator With
  Input Saturation}.
\newblock {\em IEEE Transactions on Systems, Man, and Cybernetics: Systems},
  46(3):334--344, March 2016.

\bibitem{7468475}
W.~He, Z.~Yin, and C.~Sun.
\newblock {Adaptive Neural Network Control of a Marine Vessel With Constraints
  Using the Asymmetric Barrier Lyapunov Function}.
\newblock {\em IEEE Transactions on Cybernetics}, 47(7):1641--1651, July 2017.

\bibitem{6651788}
W.~He, S.~Zhang, and S.~S. Ge.
\newblock {Adaptive Control of a Flexible Crane System With the Boundary Output
  Constraint}.
\newblock {\em IEEE Transactions on Industrial Electronics}, 61(8):4126--4133,
  Aug 2014.

\bibitem{He2015}
Wei He, Shuzhi~Sam Ge, Yanan Li, Effie Chew, and Yee~Sien Ng.
\newblock {Neural Network Control of a Rehabilitation Robot by State and Output
  Feedback}.
\newblock {\em Journal of Intelligent {\&} Robotic Systems}, 80(1):15--31, Oct
  2015.

\bibitem{HE20141843}
Wei He, Shuang Zhang, and Shuzhi~Sam Ge.
\newblock {Robust adaptive control of a thruster assisted position mooring
  system}.
\newblock {\em Automatica}, 50(7):1843 -- 1851, 2014.

\bibitem{hein2017formal}
Matthias Hein and Maksym Andriushchenko.
\newblock {Formal Guarantees on the Robustness of a Classifier against
  Adversarial Manipulation}.
\newblock In {\em Advances in Neural Information Processing Systems}, pages
  2263--2273, 2017.

\bibitem{HerbertFaSTrack2017}
Sylvia~L. Herbert, Mo~Chen, SooJean Han, Somil Bansal, Jaime~F. Fisac, and
  Claire~J. Tomlin.
\newblock {FaSTrack: a Modular Framework for Fast and Guaranteed Safe Motion
  Planning}.
\newblock {\em CoRR}, abs/1703.07373, 2017.

\bibitem{MPT3}
M.~Herceg, M.~Kvasnica, C.N. Jones, and M.~Morari.
\newblock {Multi-Parametric Toolbox 3.0}.
\newblock In {\em Proceedings of the European Control Conference}, pages
  502--510, Z\"urich, Switzerland, July 17--19 2013.
\newblock \url{http://control.ee.ethz.ch/~mpt}.

\bibitem{hornik1989multilayer}
K.~Hornik, M.~Stinchcombe, and H.~White.
\newblock {Multilayer Feedforward Networks are Universal Approximators}.
\newblock {\em Neural Networks}, 2(5):359--366, 1989.

\bibitem{huang2017safety}
Xiaowei Huang, Marta Kwiatkowska, Sen Wang, and Min Wu.
\newblock {Safety Verification of Deep Neural Networks}.
\newblock {\em CoRR}, abs/1610.06940, 2016.

\bibitem{hull2002verification}
Jason Hull, David Ward, and Radoslaw~R Zakrzewski.
\newblock {Verification and Validation of Neural Networks for Safety-Critical
  Applications}.
\newblock In {\em American Control Conference, 2002. Proceedings of the 2002},
  volume~6, pages 4789--4794. IEEE, 2002.

\bibitem{IegorovPeriodic2017}
Oleg Iegorov, Reinier Torres, and Sebastian Fischmeister.
\newblock {Periodic Task Mining in Embedded System Traces}.
\newblock pages 331--340, 04 2017.

\bibitem{IsbernerOpenSource2015}
Malte Isberner, Falk Howar, and Bernhard Steffen.
\newblock {The Open-Source LearnLib}.
\newblock In Daniel Kroening and Corina~S. P{\u{a}}s{\u{a}}reanu, editors, {\em
  Computer Aided Verification}, pages 487--495, Cham, 2015. Springer
  International Publishing.

\bibitem{JhaBoolean2017}
S.~Jha, V.~Raman, and T.and Francis~M. Pinto, A.and~Sahai.
\newblock {On Learning Sparse Boolean Formulae for Explaining AI Decisions}.
\newblock In {\em NFM 2017: NASA Formal Methods}, pages 99--114. Springer,
  2017.

\bibitem{JhaTeLEx2017}
S.~Jha, A.~Tiwari, and T.and Shankar~N. Seshia, S.A.and~Sahai.
\newblock {TeLEx: Passive STL Learning Using Only Positive Examples}.
\newblock In {\em RV 2017: Runtime Verification}, pages 208--224. Springer,
  2017.

\bibitem{JI20177158}
Huihui Ji, He~Zhang, and Tian Senping.
\newblock {Reachable set estimation for inertial Markov jump BAM neural network
  with partially unknown transition rates and bounded disturbances}.
\newblock {\em Journal of the Franklin Institute}, 354(15):7158 -- 7182, 2017.

\bibitem{jin2015mining}
Xiaoqing Jin, Alexandre Donz{\'e}, Jyotirmoy~V Deshmukh, and Sanjit~A Seshia.
\newblock Mining requirements from closed-loop control models.
\newblock {\em IEEE Transactions on Computer-Aided Design of Integrated
  Circuits and Systems}, 34(11):1704--1717, 2015.

\bibitem{julian2016policy}
K.~Julian, J.~Lopez, J.~Brush, M.~Owen, and M.~Kochenderfer.
\newblock {Policy Compression for Aircraft Collision Avoidance Systems}.
\newblock In {\em 35th Digital Avionics Systems Conference. (DASC)}, pages
  1--10, 2016.

\bibitem{julian2017neural}
Kyle Julian and Mykel~J. Kochenderfer.
\newblock {Neural Network Guidance for UAVs}.
\newblock In {\em AIAA Guidance Navigation and Control Conference (GNC)}, 2017.

\bibitem{KaelblingSurvey1996}
Leslie~Pack Kaelbling, Michael~L. Littman, and Andrew~W. Moore.
\newblock {Reinforcement Learning: A Survey}.
\newblock {\em CoRR}, cs.AI/9605103, 1996.

\bibitem{kamalapurkar2015approximate}
Rushikesh Kamalapurkar, Huyen Dinh, Shubhendu Bhasin, and Warren~E Dixon.
\newblock Approximate optimal trajectory tracking for continuous-time nonlinear
  systems.
\newblock {\em Automatica}, 51:40--48, 2015.

\bibitem{kamalapurkar2016efficient}
Rushikesh Kamalapurkar, Joel~A Rosenfeld, and Warren~E Dixon.
\newblock Efficient model-based reinforcement learning for approximate online
  optimal control.
\newblock {\em Automatica}, 74:247--258, 2016.

\bibitem{Kamalapurkar2013ConcurrentLA}
Rushikesh Kamalapurkar, Patrick Walters, and Warren~E. Dixon.
\newblock Concurrent learning-based approximate optimal regulation.
\newblock {\em 52nd IEEE Conference on Decision and Control}, pages 6256--6261,
  2013.

\bibitem{kamalapurkar2018reinforcement}
Rushikesh Kamalapurkar, Patrick Walters, Joel Rosenfeld, and Warren Dixon.
\newblock {\em Reinforcement Learning for Optimal Feedback Control: A
  Lyapunov-Based Approach}.
\newblock Springer, 2018.

\bibitem{katz2017reluplex}
G.~Katz, C.~Barrett, D.~Dill, K.~Julian, and M.~Kochenderfer.
\newblock {Reluplex: An Efficient SMT Solver for Verifying Deep Neural
  Networks}.
\newblock In {\em International Conference on Computer Aided Verification},
  pages 97--117. Springer, 2017.

\bibitem{363441}
Chao-Chee Ku and K.~Y. Lee.
\newblock {Diagonal Recurrent Neural Networks for Dynamic Systems Control}.
\newblock {\em IEEE Transactions on Neural Networks}, 6(1):144--156, Jan 1995.

\bibitem{KumarAdversarial2017}
Atul Kumar, Sameep Mehta, and Deepak Vijaykeerthy.
\newblock {An Introduction to Adversarial Machine Learning}.
\newblock In P.~Krishna Reddy, Ashish Sureka, Sharma Chakravarthy, and Subhash
  Bhalla, editors, {\em Big Data Analytics}, pages 293--299, Cham, 2017.
  Springer International Publishing.

\bibitem{10.1007/978-3-540-30138-7_3}
Zeshan Kurd and Tim~P. Kelly.
\newblock {Using Fuzzy Self-Organising Maps for Safety Critical Systems}.
\newblock In Maritta Heisel, Peter Liggesmeyer, and Stefan Wittmann, editors,
  {\em Computer Safety, Reliability, and Security}, pages 17--30, Berlin,
  Heidelberg, 2004. Springer Berlin Heidelberg.

\bibitem{LangleyML2011}
Pat Langley.
\newblock {The Changing Science of Machine Learning}.
\newblock {\em Machine Learning}, 82(3):275--279, Mar 2011.

\bibitem{Leofante2018}
F.~{Leofante}, N.~{Narodytska}, L.~{Pulina}, and A.~{Tacchella}.
\newblock {Automated Verification of Neural Networks: Advances, Challenges and
  Perspectives}.
\newblock {\em ArXiv e-prints}, May 2018.

\bibitem{Lewis}
F.L. Lewis and Shuzhi Sam~Ge.
\newblock {Neural Networks in Feedback Control Systems}.
\newblock pages 791 -- 825, 02 2006.

\bibitem{lewis2012optimal}
Frank~L Lewis, Draguna Vrabie, and Vassilis~L Syrmos.
\newblock {\em Optimal control}.
\newblock John Wiley \& Sons, 2012.

\bibitem{QingkaiAcceleratedSafeRL2018}
Qingkai Liang, Fanyu Que, and Eytan Modiano.
\newblock {Accelerated Primal-Dual Policy Optimization for Safe Reinforcement
  Learning}.
\newblock {\em CoRR}, abs/1802.06480, 2018.

\bibitem{liberzon2011calculus}
Daniel Liberzon.
\newblock {\em Calculus of variations and optimal control theory: a concise
  introduction}.
\newblock Princeton University Press, 2011.

\bibitem{7429795}
Y.~J. Liu, J.~Li, S.~Tong, and C.~L.~P. Chen.
\newblock {Neural Network Control-Based Adaptive Learning Design for Nonlinear
  Systems With Full-State Constraints}.
\newblock {\em IEEE Transactions on Neural Networks and Learning Systems},
  27(7):1562--1571, July 2016.

\bibitem{Lomuscio2017an_arxiv}
Alessio Lomuscio and Lalit Maganti.
\newblock {An approach to reachability analysis for feed-forward ReLU neural
  networks}.
\newblock {\em CoRR}, abs/1706.07351, 2017.

\bibitem{LeiMaDeepGauge2018}
Lei Ma, Felix Juefei{-}Xu, Jiyuan Sun, Chunyang Chen, Ting Su, Fuyuan Zhang,
  Minhui Xue, Bo~Li, Li~Li, Yang Liu, Jianjun Zhao, and Yadong Wang.
\newblock {DeepGauge: Comprehensive and Multi-Granularity Testing Criteria for
  Gauging the Robustness of Deep Learning Systems}.
\newblock {\em CoRR}, abs/1803.07519, 2018.

\bibitem{MedhatFramework2015}
Ramy Medhat, S.~Ramesh, Borzoo Bonakdarpour, and Sebastian Fischmeister.
\newblock {A Framework for Mining Hybrid Automata from Input/Output Traces}.
\newblock In {\em Proceedings of the 12th International Conference on Embedded
  Software}, EMSOFT '15, pages 177--186, Piscataway, NJ, USA, 2015. IEEE Press.

\bibitem{MertenNextGeneration2011}
Maik Merten, Bernhard Steffen, Falk Howar, and Tiziana Margaria.
\newblock {Next Generation LearnLib}.
\newblock In Parosh~Aziz Abdulla and K.~Rustan~M. Leino, editors, {\em Tools
  and Algorithms for the Construction and Analysis of Systems}, pages 220--223,
  Berlin, Heidelberg, 2011. Springer Berlin Heidelberg.

\bibitem{MunosSafeEfficientRL2016}
R{\'{e}}mi Munos, Tom Stepleton, Anna Harutyunyan, and Marc~G. Bellemare.
\newblock {Safe and Efficient Off-Policy Reinforcement Learning}.
\newblock {\em CoRR}, abs/1606.02647, 2016.

\bibitem{572089}
K.~S. Narendra and S.~Mukhopadhyay.
\newblock {Adaptive Control Using Neural Networks and Approximate Models}.
\newblock {\em IEEE Transactions on Neural Networks}, 8(3):475--485, May 1997.

\bibitem{80202}
K.~S. Narendra and K.~Parthasarathy.
\newblock {Identification and Control of Dynamical Systems Using Neural
  Networks}.
\newblock {\em IEEE Transactions on Neural Networks}, 1(1):4--27, Mar 1990.

\bibitem{narodytska2017verifying}
Nina Narodytska, Shiva~Prasad Kasiviswanathan, Leonid Ryzhyk, Mooly Sagiv, and
  Toby Walsh.
\newblock {Verifying Properties of Binarized Deep Neural Networks}.
\newblock {\em arXiv preprint arXiv:1709.06662}, 2017.

\bibitem{NiggemannLearning2012}
Oliver Niggemann, Benno Stein, Asmir Vodencarevic, Alexander~G Maier, and
  Hans~Kleine B{\"u}ning.
\newblock {Learning Behavior Models for Hybrid Timed Systems}.
\newblock In {\em AAAI}, 2012.

\bibitem{novak2018sensitivity}
R.~{Novak}, Y.~{Bahri}, D.~A. {Abolafia}, J.~{Pennington}, and
  J.~{Sohl-Dickstein}.
\newblock {Sensitivity and Generalization in Neural Networks: an Empirical
  Study}.
\newblock {\em ArXiv e-prints}, February 2018.

\bibitem{OlssonActive2009}
Fredrik Olsson.
\newblock {\em {A literature survey of active machine learning in the context
  of natural language processing}}.
\newblock SICS Technical Report. 1 edition, 2009.

\bibitem{PathakSafeRLVerification2018}
Shashank Pathak, Luca Pulina, and Armando Tacchella.
\newblock {Verification and repair of control policies for safe reinforcement
  learning}.
\newblock {\em Applied Intelligence}, 48(4):886--908, Apr 2018.

\bibitem{pei2017deepxplore}
Kexin Pei, Yinzhi Cao, Junfeng Yang, and Suman Jana.
\newblock {DeepXplore: Automated Whitebox Testing of Deep Learning Systems}.
\newblock {\em CoRR}, abs/1705.06640, 2017.

\bibitem{PerkinsLyapunov}
Theodore~J. Perkins and Andrew~G. Barto.
\newblock {Lyapunov Design for Safe Reinforcement Learning}.
\newblock {\em J. Mach. Learn. Res.}, 3:803--832, March 2003.

\bibitem{PintoRobust}
Lerrel Pinto, James Davidson, Rahul Sukthankar, and Abhinav Gupta.
\newblock {Robust Adversarial Reinforcement Learning}.
\newblock {\em CoRR}, abs/1703.02702, 2017.

\bibitem{pulina2010abstraction}
Luca Pulina and Armando Tacchella.
\newblock {An Abstraction-Refinement Approach to Verification of Artificial
  Neural Networks}.
\newblock In Tayssir Touili, Byron Cook, and Paul Jackson, editors, {\em
  Computer Aided Verification}, pages 243--257, Berlin, Heidelberg, 2010.
  Springer Berlin Heidelberg.

\bibitem{PulinaNever}
Luca Pulina and Armando Tacchella.
\newblock {NeVer: a tool for artificial neural networks verification}.
\newblock {\em Annals of Mathematics and Artificial Intelligence},
  62(3):403--425, Jul 2011.

\bibitem{pulina2012challenging}
Luca Pulina and Armando Tacchella.
\newblock {Challenging SMT solvers to verify neural networks}.
\newblock {\em AI Communications}, 25(2):117--135, 2012.

\bibitem{RaffeltLearnLib2005}
Harald Raffelt, Bernhard Steffen, and Therese Berg.
\newblock {LearnLib: A Library for Automata Learning and Experimentation}.
\newblock In {\em Proceedings of the 10th International Workshop on Formal
  Methods for Industrial Critical Systems}, FMICS '05, pages 62--71, New York,
  NY, USA, 2005. ACM.

\bibitem{raghunathan2018certified}
Aditi Raghunathan, Jacob Steinhardt, and Percy Liang.
\newblock {Certified Defenses against Adversarial Examples}.
\newblock {\em CoRR}, abs/1801.09344, 2018.

\bibitem{RuanDeepGO2018}
W.~{Ruan}, X.~{Huang}, and M.~{Kwiatkowska}.
\newblock {Reachability Analysis of Deep Neural Networks with Provable
  Guarantees}.
\newblock {\em ArXiv e-prints}, May 2018.

\bibitem{ruan2018global}
W.~{Ruan}, M.~{Wu}, Y.~{Sun}, X.~{Huang}, D.~{Kroening}, and M.~{Kwiatkowska}.
\newblock {Global Robustness Evaluation of Deep Neural Networks with Provable
  Guarantees for $L_0$ Norm}.
\newblock {\em ArXiv e-prints}, April 2018.

\bibitem{7086072}
A.~Sahoo, H.~Xu, and S.~Jagannathan.
\newblock {Neural Network-Based Event-Triggered State Feedback Control of
  Nonlinear Continuous-Time Systems}.
\newblock {\em IEEE Transactions on Neural Networks and Learning Systems},
  27(3):497--509, March 2016.

\bibitem{165588}
R.~M. Sanner and J.~J.~E. Slotine.
\newblock {Gaussian Networks for Direct Adaptive Control}.
\newblock {\em IEEE Transactions on Neural Networks}, 3(6):837--863, Nov 1992.

\bibitem{SaundersTrial2017}
William Saunders, Girish Sastry, Andreas Stuhlm{\"{u}}ller, and Owain Evans.
\newblock {Trial without Error: Towards Safe Reinforcement Learning via Human
  Intervention}.
\newblock {\em CoRR}, abs/1707.05173, 2017.

\bibitem{scheibler2015towards}
Karsten Scheibler, Leonore Winterer, Ralf Wimmer, and Bernd Becker.
\newblock {Towards Verification of Artificial Neural Networks.}
\newblock In {\em MBMV}, pages 30--40, 2015.

\bibitem{SettlesActiveLearning2010}
Burr Settles.
\newblock Active learning literature survey.
\newblock Computer Sciences Technical Report 1648, University of
  Wisconsin--Madison, 2009.

\bibitem{ShalevMultiRL}
Shai Shalev{-}Shwartz, Shaked Shammah, and Amnon Shashua.
\newblock {Safe, Multi-Agent, Reinforcement Learning for Autonomous Driving}.
\newblock {\em CoRR}, abs/1610.03295, 2016.

\bibitem{MuddassarIDS2012}
Muddassar~A. Sindhu and Karl Meinke.
\newblock {IDS: An Incremental Learning Algorithm for Finite Automata}.
\newblock {\em CoRR}, abs/1206.2691, 2012.

\bibitem{SinhaCertifying2017}
A.~{Sinha}, H.~{Namkoong}, and J.~{Duchi}.
\newblock {Certifying Some Distributional Robustness with Principled
  Adversarial Training}.
\newblock {\em ArXiv e-prints}, October 2017.

\bibitem{SmeenkApplying2015}
Wouter Smeenk, Joshua Moerman, Frits Vaandrager, and David~N. Jansen.
\newblock {Applying Automata Learning to Embedded Control Software}.
\newblock In Michael Butler, Sylvain Conchon, and Fatiha Za{\"i}di, editors,
  {\em Formal Methods and Software Engineering}, pages 67--83, Cham, 2015.
  Springer International Publishing.

\bibitem{stingu2011approximate}
Emanuel Stingu and Frank~L Lewis.
\newblock An approximate dynamic programming based controller for an
  underactuated 6dof quadrotor.
\newblock In {\em Adaptive Dynamic Programming And Reinforcement Learning
  (ADPRL), 2011 IEEE Symposium on}, pages 271--278. IEEE, 2011.

\bibitem{SummervilleCHARDA2017}
Adam Summerville, Joseph~C. Osborn, and Michael Mateas.
\newblock {CHARDA: Causal Hybrid Automata Recovery via Dynamic Analysis}.
\newblock {\em CoRR}, abs/1707.03336, 2017.

\bibitem{SunConcolic2018}
Y.~{Sun}, M.~{Wu}, W.~{Ruan}, X.~{Huang}, M.~{Kwiatkowska}, and D.~{Kroening}.
\newblock {Concolic Testing for Deep Neural Networks}.
\newblock {\em ArXiv e-prints}, April 2018.

\bibitem{sun2018testing}
Youcheng Sun, Xiaowei Huang, and Daniel Kroening.
\newblock {Testing Deep Neural Networks}.
\newblock {\em arXiv preprint arXiv:1803.04792}, 2018.

\bibitem{szegedy2013intriguing}
Christian Szegedy, Wojciech Zaremba, Ilya Sutskever, Joan Bruna, Dumitru Erhan,
  Ian~J. Goodfellow, and Rob Fergus.
\newblock {Intriguing properties of neural networks}.
\newblock {\em CoRR}, abs/1312.6199, 2013.

\bibitem{501721}
K.~Tanaka.
\newblock {An Approach to Stability Criteria of Neural-Network Control
  Systems}.
\newblock {\em IEEE Transactions on Neural Networks}, 7(3):629--642, May 1996.

\bibitem{ThomasSafeRL}
Philip Thomas.
\newblock {\em {Safe Reinforcement Learning}}.
\newblock PhD thesis, University of Massachusetts Amherst, 2015.

\bibitem{tian2017deeptest}
Yuchi Tian, Kexin Pei, Suman Jana, and Baishakhi Ray.
\newblock {DeepTest: Automated Testing of Deep-Neural-Network-driven Autonomous
  Cars}.
\newblock {\em CoRR}, abs/1708.08559, 2017.

\bibitem{tjeng2017verifying}
Vincent Tjeng and Russ Tedrake.
\newblock {Verifying Neural Networks with Mixed Integer Programming}.
\newblock {\em arXiv preprint arXiv:1711.07356}, 2017.

\bibitem{2018arXiv180406760E}
C.~Erkan {Tuncali}, G.~{Fainekos}, H.~{Ito}, and J.~{Kapinski}.
\newblock {Simulation-based Adversarial Test Generation for Autonomous Vehicles
  with Machine Learning Components}.
\newblock {\em ArXiv e-prints}, April 2018.

\bibitem{DBLP:journals/corr/abs-1804-03973}
Cumhur~Erkan Tuncali, James Kapinski, Hisahiro Ito, and Jyotirmoy~V. Deshmukh.
\newblock {Reasoning about Safety of Learning-Enabled Components in Autonomous
  Cyber-physical Systems}.
\newblock {\em CoRR}, abs/1804.03973, 2018.

\bibitem{vamvoudakis2010online}
Kyriakos~G Vamvoudakis and Frank~L Lewis.
\newblock Online actor--critic algorithm to solve the continuous-time infinite
  horizon optimal control problem.
\newblock {\em Automatica}, 46(5):878--888, 2010.

\bibitem{Robert2001linear}
Robert~J. Vanderbei.
\newblock {\em {Linear Programming: Foundations \& Extensions (Second
  Edition).}}
\newblock Springer, 2001.

\bibitem{VerwerEfficient2010}
S.~E. Verwer.
\newblock {\em {Efficient Identification of Timed Automata: Theory and
  Practice}}.
\newblock {PhD} dissertation, Delft University of Technology, 2010.

\bibitem{VerwerAlgorithm2007}
Sicco Verwer, Mathijs de~Weerdt, and Cees Witteveen.
\newblock {\em {An algorithm for learning real-time automata}}.
\newblock 2007.

\bibitem{VerwerEfficiency2011}
Sicco Verwer, Mathijs de~Weerdt, and Cees Witteveen.
\newblock {The efficiency of identifying timed automata and the power of
  clocks}.
\newblock {\em Information and Computation}, 209(3):606 -- 625, 2011.
\newblock Special Issue: 3rd International Conference on Language and Automata
  Theory and Applications (LATA 2009).

\bibitem{VerwerEfficiently2012}
Sicco Verwer, Mathijs de~Weerdt, and Cees Witteveen.
\newblock {Efficiently identifying deterministic real-time automata from
  labeled data}.
\newblock {\em Machine Learning}, 86(3):295--333, Mar 2012.

\bibitem{vijayakumar2018neural}
Ashwin~J Vijayakumar, Abhishek Mohta, Oleksandr Polozov, Dhruv Batra, Prateek
  Jain, and Sumit Gulwani.
\newblock {Neural-Guided Deductive Search for Real-Time Program Synthesis from
  Examples}.
\newblock {\em arXiv preprint arXiv:1804.01186}, 2018.

\bibitem{vrabie2009neural}
Draguna Vrabie and Frank Lewis.
\newblock Neural network approach to continuous-time direct adaptive optimal
  control for partially unknown nonlinear systems.
\newblock {\em Neural Networks}, 22(3):237--246, 2009.

\bibitem{6937163}
H.~Wang, K.~Liu, X.~Liu, B.~Chen, and C.~Lin.
\newblock {Neural-Based Adaptive Output-Feedback Control for a Class of
  Nonstrict-Feedback Stochastic Nonlinear Systems}.
\newblock {\em IEEE Transactions on Cybernetics}, 45(9):1977--1987, Sept 2015.

\bibitem{7066958}
H.~Wang, X.~Liu, and K.~Liu.
\newblock {Robust Adaptive Neural Tracking Control for a Class of Stochastic
  Nonlinear Interconnected Systems}.
\newblock {\em IEEE Transactions on Neural Networks and Learning Systems},
  27(3):510--523, March 2016.

\bibitem{WangFormal2018}
S.~{Wang}, K.~{Pei}, J.~{Whitehouse}, J.~{Yang}, and S.~{Jana}.
\newblock {Formal Security Analysis of Neural Networks using Symbolic
  Intervals}.
\newblock {\em ArXiv e-prints}, April 2018.

\bibitem{7087381}
T.~Wang, H.~Gao, and J.~Qiu.
\newblock {A Combined Adaptive Neural Network and Nonlinear Model Predictive
  Control for Multirate Networked Industrial Process Control}.
\newblock {\em IEEE Transactions on Neural Networks and Learning Systems},
  27(2):416--425, Feb 2016.

\bibitem{WEI2015106}
Qinglai Wei and Derong Liu.
\newblock {Neural-network-based adaptive optimal tracking control scheme for
  discrete-time nonlinear systems with approximation errors}.
\newblock {\em Neurocomputing}, 149:106 -- 115, 2015.
\newblock Advances in neural networks Advances in Extreme Learning Machines.

\bibitem{2018arXiv180110578W}
T.-W. {Weng}, H.~{Zhang}, P.-Y. {Chen}, J.~{Yi}, D.~{Su}, Y.~{Gao}, C.-J.
  {Hsieh}, and L.~{Daniel}.
\newblock {Evaluating the Robustness of Neural Networks: An Extreme Value
  Theory Approach}.
\newblock {\em ArXiv e-prints}, January 2018.

\bibitem{weng2018towards}
Tsui-Wei Weng, Huan Zhang, Hongge Chen, Zhao Song, Cho-Jui Hsieh, Duane Boning,
  Inderjit~S Dhillon, and Luca Daniel.
\newblock {Towards Fast Computation of Certified Robustness for ReLU Networks}.
\newblock {\em arXiv preprint arXiv:1804.09699}, 2018.

\bibitem{wicker2018feature}
Matthew Wicker, Xiaowei Huang, and Marta Kwiatkowska.
\newblock {Feature-Guided Black-Box Safety Testing of Deep Neural Networks}.
\newblock In {\em International Conference on Tools and Algorithms for the
  Construction and Analysis of Systems}, pages 408--426. Springer, 2018.

\bibitem{kolter2017provable}
Eric Wong and J~Zico Kolter.
\newblock {Provable Defenses against Adversarial Examples via the Convex Outer
  Adversarial Polytope}.
\newblock {\em arXiv preprint arXiv:1711.00851}, 2017.

\bibitem{xiang2018reachable}
Weiming Xiang, Diego~Manzanas Lopez, Patrick Musau, and Taylor~T Johnson.
\newblock {Reachable Set Estimation and Verification for Neural Network Models
  of Nonlinear Dynamic Systems}.
\newblock {\em arXiv preprint arXiv:1802.03557}, 2018.

\bibitem{xiang2017reachable_arxiv}
Weiming Xiang, Hoang-Dung Tran, and Taylor~T Johnson.
\newblock {Reachable Set Computation and Safety Verification for Neural
  Networks with ReLU Activations}.
\newblock {\em arXiv preprint arXiv: 1712.08163}, 2017.

\bibitem{xiang2017output}
Weiming Xiang, Hoang-Dung Tran, and Taylor~T Johnson.
\newblock {Output Reachable Set Estimation and Verification for Multi-Layer
  Neural Networks}.
\newblock {\em IEEE Transactions on Neural Network and Learning Systems}, 2018.

\bibitem{xiang2018reachable_acc}
Weiming Xiang, Hoang-Dung Tran, Joel~A Rosenfeld, and Taylor~T Johnson.
\newblock {Reachable Set Estimation and Safety Verification for Piecewise
  Linear Systems with Neural Network Controllers}.
\newblock {\em arXiv preprint arXiv:1802.06981}, 2018.

\bibitem{XieFuzzing2018}
X.~{Xie}, L.~{Ma}, F.~{Juefei-Xu}, H.~{Chen}, M.~{Xue}, B.~{Li}, Y.~{Liu},
  J.~{Zhao}, J.~{Yin}, and S.~{See}.
\newblock {Coverage-Guided Fuzzing for Deep Neural Networks}.
\newblock {\em ArXiv e-prints}, September 2018.

\bibitem{6783745}
B.~Xu, Z.~Shi, C.~Yang, and F.~Sun.
\newblock {Composite Neural Dynamic Surface Control of a Class of Uncertain
  Nonlinear Systems in Strict-Feedback Form}.
\newblock {\em IEEE Transactions on Cybernetics}, 44(12):2626--2634, Dec 2014.

\bibitem{Xu2015}
Bin Xu.
\newblock {Robust adaptive neural control of flexible hypersonic flight vehicle
  with dead-zone input nonlinearity}.
\newblock {\em Nonlinear Dynamics}, 80(3):1509--1520, May 2015.

\bibitem{ZakrzewskiRandomized2004}
R.~R. Zakrzewski.
\newblock {Randomized Approach to Verification of Neural Networks}.
\newblock In {\em 2004 IEEE International Joint Conference on Neural Networks
  (IEEE Cat. No.04CH37541)}, volume~4, pages 2819--2824 vol.4, July 2004.

\bibitem{6736141}
H.~Zhang, C.~Qin, and Y.~Luo.
\newblock {Neural-Network-Based Constrained Optimal Control Scheme for
  Discrete-Time Switched Nonlinear System Using Dual Heuristic Programming}.
\newblock {\em IEEE Transactions on Automation Science and Engineering},
  11(3):839--849, July 2014.

\bibitem{ZhangCar2017}
Yihuan Zhang, Qin Lin, Jun Wang, and Sicco Verwer.
\newblock {Car-following Behavior Model Learning Using Timed Automata}.
\newblock {\em IFAC-PapersOnLine}, 50(1):2353 -- 2358, 2017.
\newblock 20th IFAC World Congress.

\bibitem{7185423}
X.~Zhao, P.~Shi, X.~Zheng, and J.~Zhang.
\newblock {Intelligent Tracking Control for a Class of Uncertain High-Order
  Nonlinear Systems}.
\newblock {\em IEEE Transactions on Neural Networks and Learning Systems},
  27(9):1976--1982, Sept 2016.

\bibitem{ZHAO2015193}
Xudong Zhao, Peng Shi, Xiaolong Zheng, and Lixian Zhang.
\newblock {Adaptive tracking control for switched stochastic nonlinear systems
  with unknown actuator dead-zone}.
\newblock {\em Automatica}, 60:193 -- 200, 2015.

\bibitem{1296692}
Quanmin Zhu and Lingzhong Guo.
\newblock {Stable Adaptive Neurocontrol for Nonlinear Discrete-Time Systems}.
\newblock {\em IEEE Transactions on Neural Networks}, 15(3):653--662, May 2004.

\bibitem{DBLP:journals/corr/ZribiCD15}
Ali Zribi, Mohamed Chtourou, and Mohamed Djemel.
\newblock {A New PID Neural Network Controller Design for Nonlinear Processes}.
\newblock {\em CoRR}, abs/1512.07529, 2015.

\end{thebibliography}
